\documentclass[preprint,12pt]{elsarticle}

\usepackage{amssymb}

\journal{Computers in Industry}

\usepackage{graphicx}
\usepackage{amsmath}
\usepackage{xcolor}
\usepackage{multirow}
\usepackage{soul}
\usepackage{natbib}
\usepackage{url}
\usepackage{array}
\usepackage{booktabs}

\begin{document}

\begin{frontmatter}



\title{A DCNN-based Arbitrarily-Oriented Object Detector with Application to Quality Control and Inspection}


\author{Kai Yao, Alberto Ortiz*, Francisco Bonnin-Pascual}
\address{Department of Mathematics and Computer Science (University of the Balearic Islands) and IdiSBa (Health Research Institute of the Balearic Islands),
            Cra. de Valldemossa km7.5, 
            Palma de Mallorca,
            07122, 
            Balearic Islands,
            Spain}
\cortext[cor1]{Corresponding author. E-mail address: alberto.ortiz@uib.es (A. Ortiz).}            

%

\begin{abstract}
Following the successful contribution of machine vision systems to automated inspection and quality control, in this paper, we propose a new bounding boxes-based regression solution aiming at recognizing generic targets for which detecting their orientation may be beneficial. Our solution consists in a two-stage arbitrarily-oriented object detection method making use of indirect regression of oriented bounding boxes parameters. Besides, in order to be able to recognize targets of different sizes and shapes, the solution adopts a multi-scale approach. The resulting performance is evaluated against datasets from two different industry-related case studies: while one involves the detection of a number of object classes in the context of a quality control application, the other stems from the visual inspection domain and deals with the localization of image areas corresponding to scene points affected by a specific sort of defect. The detection results that are reported for both tasks show that competitive performance levels can be achieved in both cases despite the differences among them and their specific challenges.
\end{abstract}



\begin{keyword}
Quality Control and Inspection \sep Deep Learning \sep Object Recognition \sep Bounding Boxes Regression
\end{keyword}

\end{frontmatter}




\section{Introduction}

In contrast to human-based inspection and monitoring, automatic inspection techniques have the advantage of high efficiency, low cost and objectivity~\cite{Gao2021}. In this regard, machine vision technologies are emerging as more and more popular solutions for this sort of tasks, as they enable non-contact, thus non-destructive, inspection procedures. As a consequence, along the previous years, a plethora of industrial activities have benefited from the incorporation of machine vision-based systems into the underlying processes, encompassing an increasingly wide range of industrial sectors such as steel~\cite{Luo2019}, wood~\cite{Kamal2017}, ceramic~\cite{Zhao2021}, fabric~\cite{Cao2017}, industrial supplies~\cite{Leo2017}, electronic circuits/components~\cite{Liu2021} and food~\cite{Zhu2021}, as well as infrastructure surveying~\cite{Chen2021b}, to name but a few. 

In the past several years, many vision-based methods have been proposed, and some newly-emerged techniques, such as deep learning, have become increasingly popular and have addressed many challenging problems effectively~\cite{Bertolini2021,Chai2021}. Regarding the latter, Deep Convolutional Neural Networks (DCNN) have demonstrated remarkable capabilities for problems so complex as image classification, multi-instance multi-object detection or multi-class semantic segmentation on the basis of the CNN concept proposed by LeCun and his collaborators (leading to the well-known LeNet networks~\cite{LeCun1998}), followed by the technological breaktrough that allowed training artificial neural structures with a number of parameters amounting to millions~\cite{Krizhevsky2012}. This breakthrough in performance is commonly established as the consequence of the \textit{learning the representation} capacity of DCNNs, embedded in the set of multi-scale feature maps defined in their architecture through non-linear activation functions and a hierarchy of convolutional filters that are automatically learnt during the training process by means of iterative back-propagation of prediction errors between current and expected output. As a result, and in contrast to manually designed image processing solutions, DCNNs automatically generate powerful features from training data.

\begin{figure}
\centering
\begin{tabular}{cc}
\includegraphics[height=4cm]{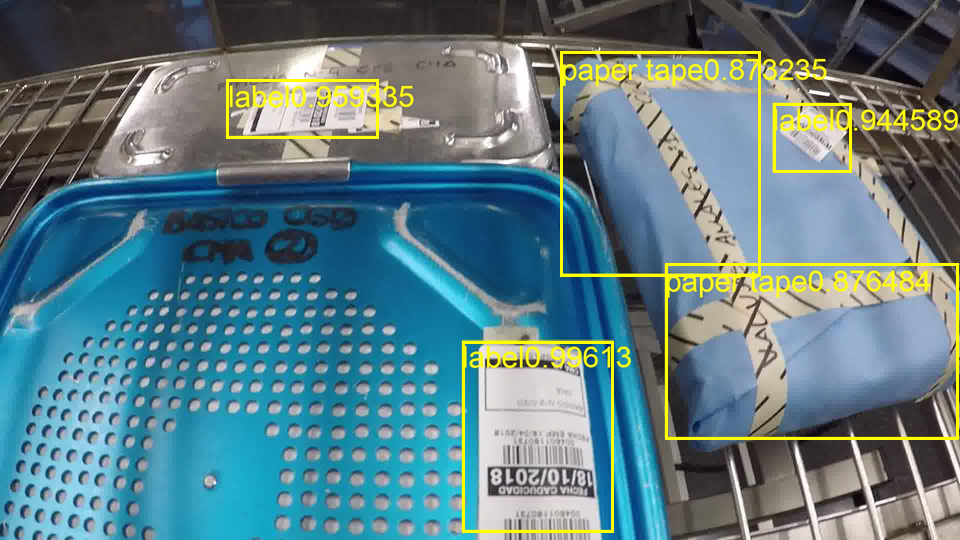} &
\includegraphics[height=4cm]{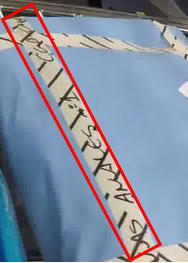}
\end{tabular}
\caption{Use of oriented bounding boxes for objects with different shapes, sizes and aspect ratios: (left) example of detection by means of unoriented bounding boxes, (right) example of a more effective detection by means of an oriented bounding box for the OI with the largest aspect ratio in the left image.}
\label{fig:bboxes}
\end{figure}

In this work, we adopt DCNN-based methodologies with an orientation towards multi-class object recognition, a domain for which DCNNs have shown very competitive performance under different operating conditions and with a minimum of human interaction or expert process knowledge~\cite{Simonyan2014a}. Our proposal is a generic solution for multi-scale, arbitrarily-oriented object detection that can be applied to any context (after proper training). By \textit{arbitrarily-oriented object detection} we mean that the output of the detector is a collection of oriented bounding boxes likely containing any of the \textit{objects of interest} (OI) for the task at hand. The fact that the detector is aware of objects orientation permits adapting the detection to the area where the object lies without involving more pixels from the background than necessary, thus producing a more effective detection (see Fig.~\ref{fig:bboxes} for an illustration). On the other side, a multi-scale detector allows dealing with objects that can appear in different sizes. Both features, multi-scale and oriented-detection, become necessary when the detection problem involves objects with very different shapes, sizes and aspect ratios, such as the ones involved in the two detection problems that have motivated this work, stemming from the quality control and visual inspection domains. They both constitute application areas where DCNNs have also exhibited highly competitive performance~\cite{Weimer2016}.

In more detail, this paper proposes a two-stage arbitrarily-oriented detector: the first stage predicts locations for the objects of interest in the form of unoriented bounding boxes, adopting a feature pyramid-based approach to produce detections at different scales and so capture minor details if needed; the second stage implements a lightweight CNN that is used for regressing the parameters of the oriented bounding boxes better fitting the OI that lie inside the unoriented predictions produced by the first stage. 
A preliminary version of this work can be found in~\cite{Yao2019} as a work-in progress paper.


The main contributions of this work are as follows: (a) we design a two-stage arbitrarily-oriented multi-category object detector, which we show it can successfully operate in dense and complicated scenes; (b) we propose a feature pyramid-based network architecture {for multi-scale unoriented boxes regression and we also analyze several map fusion strategies across scales in order to choose the one leading to highest performance}; (c) the unoriented boxes regressor adopts a default boxes-based scheme {using the output of a clustering process on the training data to obtain data-driven high-quality priors} and improve target localization accuracy; (d) oriented bounding boxes regression is achieved by means of {an extremely light network and adopting a simple parameterization that avoids boundary problems with e.g. angle-like regression}; (e) we evaluate the performance of the full detector {in the context of two industry-related applications such that the overall set of targets to be detected exhibit important differences in scale, shape regularity and aspect ratio, and hence represent a relevant object detection benchmark; and (f) the evaluation performed includes comparative studies on some important design choices, together with comparisons with similar state-of-the-art solutions}.

The rest of the paper is organized as follows: Section~\ref{sc:related} reviews related work; Section~\ref{sc:scenarios} describes the two application scenarios we use as a benchmark for our approach; Section~\ref{sec:overview} overviews the full network, while Section~\ref{sec:fpssd} describes the multi-scale, orientation-unaware detector, Section~\ref{sec:default_boxes} outlines the default boxes selection process and Section~\ref{sec:rbox_regression} details the network producing oriented bounding boxes; Section~\ref{sc:results} reports on the results of a number of experiments aiming at showing the performance of the full detector; and Section~\ref{sc:conclusions} concludes the paper.

\section{Related Work}
\label{sc:related}

{Object detection refers to a fundamental problem in computer vision, aiming at the design of algorithms able to recognize and locate objects in pictures without human intervention. With the advent of DCNN-based methodologies, a number of objecte detection approaches have emerged in a number of domains, including remote sensing~\cite{Zou2020,Azimi2021,Sun2022}, text recognition~\cite{Liu2018,Liao2018,Xu2019,Wang2021}, face detection~\cite{Shi2018,Li2019,Zhang2020}, or surveillance~\cite{Mhalla2019,Zhou2021,Wang2022}, to name but a few.} 

{DCNN-based object detectors can be broadly classified as single- and two-stage methods~\cite{Zaidi2022}, depending on whether the solution involves a separate module to generate region proposals or not. \textit{Two-stage models} find an arbitrary number of object proposals during a first stage, and next classify them in a second stage, what leads to solutions that solve the object detection task as a classification problem~\cite{girshick2014rich,girshick2015fast,ren2015faster,Yang2019,Ding2019,Xie2021,Xu2021}. Instead, \textit{single-stage detectors} localize and classify image regions containing objects in a single shot~\cite{Liu2016, redmon2016you, Redmon2017, law2018cornernet, zhou2019objects, Pan2020, Huang2022}. Regression-based detection is the dominant approach within this family. A third class of methods named as \textit{transformer-based detectors} by obvious reasons has been recently incorporated into this broad categorization after the success of transformers in Natural Language Processing and, most importantly, in vision tasks~\cite{Liu2021,Carion2020}.}

{Most single-shot solutions adopt the so-called anchor boxes-based frameworks~\cite{Zhang2021}. \textit{Anchor boxes} are a set of predefined bounding boxes of a certain height and width. These boxes are defined to capture the scale and aspect ratio of the specific object classes of interest. During detection, the predefined anchor boxes are tiled across the image in a, usually, dense way, so that the model predicts the per-class probability and the relevant bounding box attributes at every potential position. Actually, the network does not directly predict bounding boxes, but rather refinements that correspond to the tiled anchor boxes. A threshold on the per-class probabilities and a non-maximum suppression strategy lead to the final set of predicted boxes produced by the model. The use of anchor boxes enables a network to detect at once multiple objects, objects of different scales, and overlapping objects, avoiding the scanning of the image with a sliding window.}

{A number of approaches are reviewed in the following, starting with unoriented detection methods as predecessors of the more recent orientation-aware methods.}

\subsection{Unoriented Object Detection}

{The roots of most recent and latest object detectors have to be found in those that we can consider as classic detection algorithms (in accordance to the pace at which deep learning methods are evolving), aiming at unoriented detection. Among the two-stage detectors, R-CNN~\cite{girshick2014rich} is one of the first attempts to use a two-stage neural network architecture. R-CNN regards the object detection problem as a classification task based on Regions of Interest (ROIs). In the first stage, the selective search algorithm~\cite{uijlings2013selective} is used to obtain ROIs, while, in the second stage, a DCNN is utilized to produce features that become input of a Support Vector Machine (SVM) classifier. As a consequence of the detector structure, the training process is expensive in space and time: with VGG-16~\cite{simonyan2014very}, R-CNN requires 2.5 GPU-days for the VOC2007 training set. An improved version, Fast R-CNN~\cite{girshick2015fast}, adopts an end-to-end model for object detection and significantly speeds up the training process. To be more specific, a multi-task loss function is developed to address bounding box regression and classification tasks simultaneously, and an ROI pooling layer is developed and employed to extract features after the last convolutional layer. Subsequently, Ren et al.~\cite{ren2015faster} developed the concept of Region Proposal Network (RPN) by exploiting the architecture of Fast R-CNN to generate proposals instead of the \emph{selective search} algorithm, sharing the network that produces the features. Moreover, an anchor mechanism is developed to provide prior knowledge to the bounding box regression task, making the training process more stable than before. Faster R-CNN improved detection performance and narrowed the inference time. Under the Faster R-CNN framework, the Feature Pyramid Network (FPN)~\cite{lin2017feature} employs a feature pyramid architecture to merge the feature maps from different convolutional layers by up-sampling. The multi-scale detection capability of FPN leads to better performance, especially for small objects.} 

{On the other side, some other works~\cite{Liu2016, redmon2016you, Redmon2017, law2018cornernet, zhou2019objects} formulate the object detection task as one-stage bounding box regression problems, discarding any proposal generating procedure during training. Among all of them, YOLO (You Only Look Once) is a family of computer vision models that has gained significant popularity over the years. In YOLOv1~\cite{redmon2016you}, the input picture is split into $S \times S$ grids, and each grid cell is responsible for detecting targets regardless of whether the center point of the target falls into this cell. YOLO was fast but inaccurate. To tackle down this problem, YOLOv2/YOLO9000~\cite{Redmon2017} were proposed with several improvements over the original YOLO, such as the systematic use of Batch Normalization~\cite{Ioffe2015} and the adoption of a prior boxes-based strategy for shortened training. YOLOv3~\cite{Redmon2018} increased the number of layers and anchor boxes, and made predictions at three separate levels of granularity to improve performance on smaller objects. YOLOv4 (2020) / YOLOv5 (2020), and subsequent variants that have been appearing until recently~\cite{Ganesh2022,Yao2022}, have added more or less sophisticated tweaks over previous versions to achieve enhanced performance.}

{The Single-Shot MultiBox Detector (SSD)~\cite{Liu2016} is another one-stage method of notable influence. SSD also employs a series of default boxes with different aspect ratios to help in the regression of the parameters of the bounding boxes, and adopts a simple strategy to combine multi-scale feature maps. This makes SSD superior in performance regarding multi-scale object detection. Moreover, given the usual imbalance between positive and negative samples during training, SSD follows an online hard example mining strategy (OHEM) to keep the ratio at 3:1. SSD predicts, for each anchor box, both the shape offsets $(\Delta x, \Delta y, \Delta w, \Delta h)$ and the confidences for all object categories. SSD has been shown to outperform YOLO9000 while keeping its speed and performance.}

\subsection{Arbitrarily-Oriented Object Detection}

{Over the last few years, a number of rotation-aware detectors evolved from classical detection algorithms have been developed. Consequently, the arbitrarily-oriented object detection task has also been addressed by means of single- and two-stage approaches (resp.~\cite{Pan2020,Huang2022} and~\cite{Yang2019,Ding2019,Xie2021,Xu2021}, to name but a few).} 

{In any case, box parameterization becomes a relevant aspect in respect to oriented boxes because of the consequences it may lead to. A number of variants can be found in this regard: e.g. $(x,y,w,h,\theta)$ for resp. the box center, width, height and rotation angle~\cite{Yang2019,Ding2019,Ma2018}; or $(\Delta x_1, \Delta y_1, \Delta x_2, \Delta y_2, \Delta x_3,$ $\Delta y_3, \Delta x_4, \Delta y_4)$, where the box corners are given by $p_n = (x_{0n} + w_0 \Delta x_n, y_{0n} + h_0 \Delta y_n)$, with $w_0$, $h_0$ and $(x_{0n},y_{0n})$ as resp. the width, height and coordinates of a default box~\cite{Liao2018}; or $(\delta x, \delta y, \delta w, \delta h, \delta\alpha, \delta\beta)$, where $w = a_w e^{\delta_w}$, $h = a_h e^{\delta_h}$, $x = \delta_x a_w + a_x$ and $y = \delta_y a_h + a_y$ are the width, height and center point of the external rectangle box, for a default box with parameters $(a_x,a_y,a_w,a_h)$, and $\Delta\alpha = \delta_\alpha w$, $\Delta\beta = \delta_\beta h$ are the offsets relative to the midpoints of the top and right sides of the external rectangle~\cite{Xie2021,Huang2022}; or $(x,y,w,h,\alpha_1,\alpha_2,\alpha_3,\alpha_4)$, where the box corners are given by $p_1 = (x - w/2 + \alpha_1 \cdot w, y - h/2), p_2 = (x + w/2, y - h/2 + \alpha_2 \cdot h), p_3 = (x + w/2 - \alpha_3 \cdot w, y + h/2), p_4 = (x - w/2, y + h/2 - \alpha_4 \cdot h)$~\cite{Xu2021}. In general, those approaches that adopt regression methods can be affected by discontinuous boundaries, often caused by angular periodicity (when angles are involved) or corner ordering~\cite{Yang2022}.}


{Anchor-based frameworks have also reached a significant level of maturity and have shown relevant success for oriented-boxes detection~\cite{Liao2018,Ma2018,Liu2018b,Liu2017}. For this to happen, it has been necessary to deal with a number of practical problems. In this regard,~\cite{Ding2019} uses horizontal anchors and learns the rotated boxes through spatial transformations to reduce the number of predefined anchors, while~\cite{Xu2021} develops a gliding vertex method to deal with oriented bounding boxes from unoriented boxes,~\cite{Ming2021} analyzes and proposes a dynamic matching and assignment strategy,~\cite{Xie2021} describes a two-stage solution that involves a lightweight RPN network that aims at producing oriented boxes at a very low cost which in a second stage are classified, and~\cite{Huang2022} makes use of two-dimensional oriented heatmaps based on Gaussians that are refined through a joint-optimization loss to better align the predictions with the shape and orientation of the detection targets.} 

{On the other side, to reduce the amount of computation and memory requirements of anchor-based methods, as well as to achieve larger generalization capabilities, other works~\cite{Pan2020,Duan2019,Tian2019,Wei2020,Cheng2022,Kong2020,Liu2020} propose anchor-free detectors and regress the locations directly. An alternative strategy produces oriented detections adopting an angle classification method~\cite{Yang2022,Yang2020}. To finish, a recent publication goes beyond bounding boxes regression by considering convex hull-based predictions to contemplate irregular object layouts~\cite{Guo2022}.}

{The most relevant differences of the approach that is described in this paper with regard to the previous works are as follows:
	\begin{enumerate}
		\item We propose a two-stage approach whose second stage infers a rotated box for every unoriented box regressed in the first stage, so that high-quality and already classified unoriented boxes are expected from the first stage. This justifies further the adoption of a multi-scale approach for the first stage, apart from the inherent interest on multi-scale detection.
		\item In our parameterization of oriented boxes, we do not involve angles to avoid discontinuous boundaries. Further, thanks to the two-stage approach, our second stage only needs to regress two box parameters, what considerably reduces the complexity of the network. A third parameter can be incorporated to disambiguate between the two solutions that are obtained with the two-target regression case, though our results show higher performance for the latter case.
		\item In respect to the network design methodology:
		\begin{enumerate}
		\item Our across-scale fusion approach for multi-scale detection results from the selection of one among several options in accordance to the performance attained in our benchmark.
		\item Instead of using predefined anchor boxes, we employ a clustering approach to find adequate box sizes (and by extension the corresponding aspect ratios), also in accordance to the detection targets of the intended applications.
		\end{enumerate} 
	\end{enumerate}
}

\section{Application Scenarios}
\label{sc:scenarios}

\begin{figure}[t]  
    \centering
    \begin{tabular}{@{\hspace{0mm}}c@{\hspace{1mm}}c@{\hspace{0mm}}}
        \includegraphics[height=4cm]{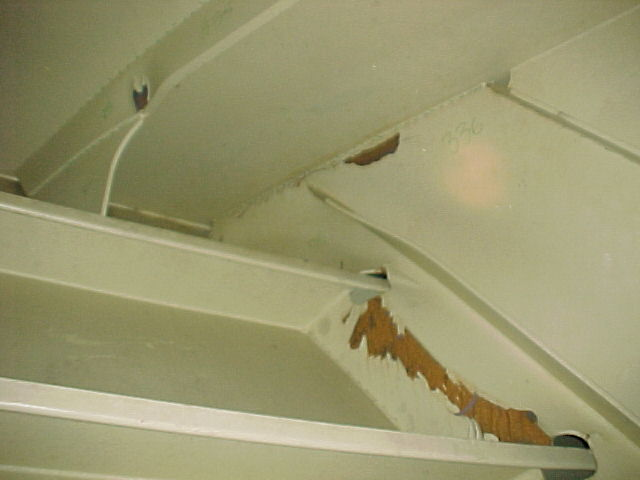}
        &
        \includegraphics[width=6.4cm,height=4cm]{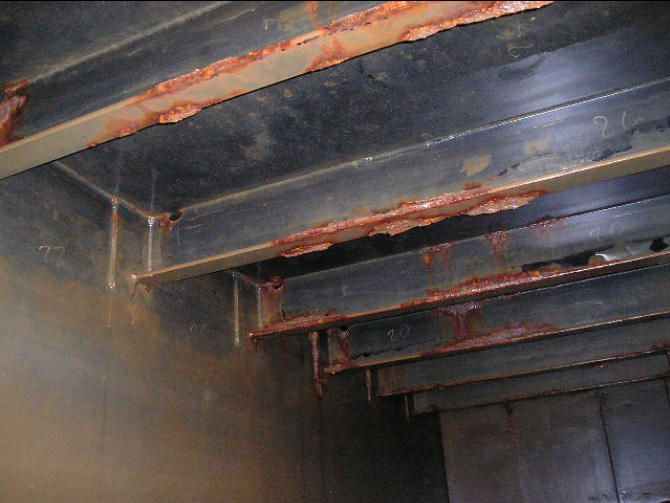}
        \\
        \includegraphics[height=4cm]{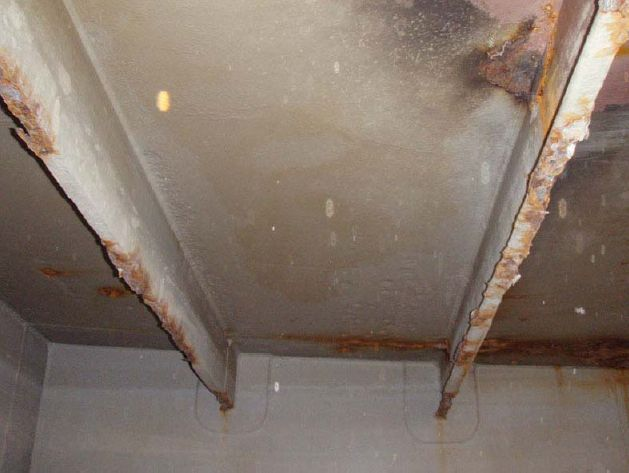}
        &
        \includegraphics[width=6.4cm,height=4cm]{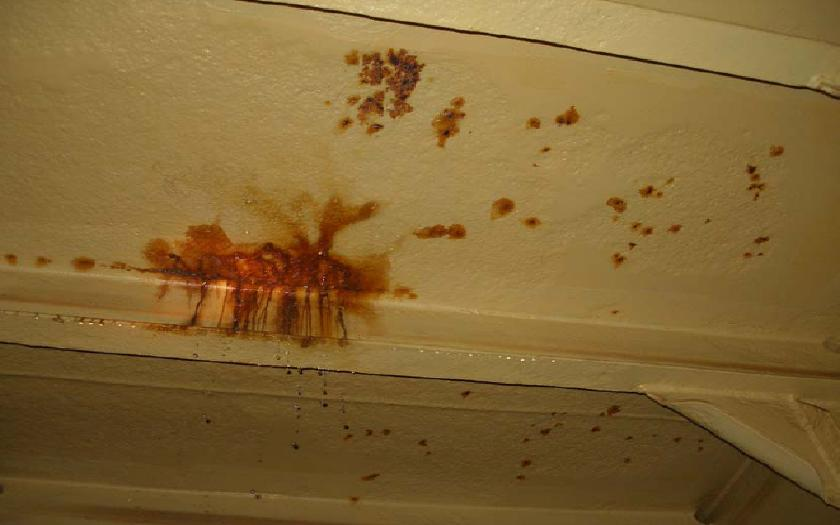}
    \end{tabular}
    \caption{Targets to be detected in the visual inspection application case. The different images show examples of CBC on ship surfaces.}
    \label{fig:targets_insp}
\end{figure}

In this work, we use two industry-related application cases as a benchmark of the object detection strategy. In the first case, we deal with the detection of one of the most common defects that can affect steel surfaces, i.e. coating breakdown and/or corrosion (CBC) in any of its many different forms. This is of particular relevance where the integrity of steel-based structures is critical, such as e.g. in large-tonnage vessels. An early detection, through suitable maintenance programmes, prevents vessel structures from suffering major damage which can ultimately compromise their integrity and lead to accidents with maybe catastrophic consequences for the crew (and passengers), environmental pollution or damage and/or total loss of the ship, its equipment and its cargo. The inspection of those ship-board structures by humans is a time-consuming, expensive and commonly hazardous activity, what, altogether, suggests the introduction of defect detection tools to alleviate the total cost of an inspection. Figure~\ref{fig:targets_insp} shows images of metallic vessel surfaces affected by CBC.

\begin{figure}[t]  
    \centering
    \begin{tabular}{@{\hspace{0mm}}c@{\hspace{1mm}}c@{\hspace{1mm}}c@{\hspace{0mm}}}
        \includegraphics[width=4cm,height=4cm]{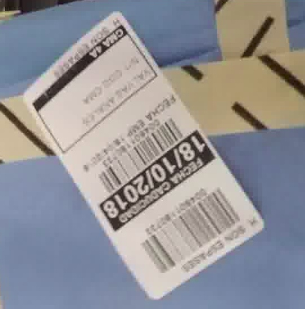}
        &
        \includegraphics[width=4cm,height=4cm]{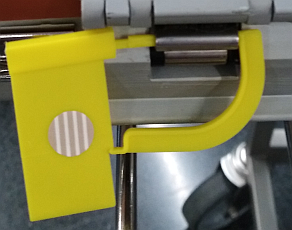}
        &
        \includegraphics[width=4cm,height=4cm]{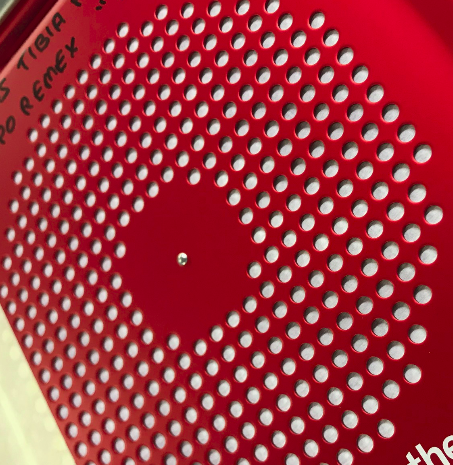} 
        \\
        \includegraphics[width=4cm,height=4cm]{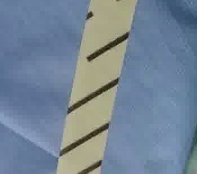} 
        &
        \includegraphics[width=4cm,height=4cm]{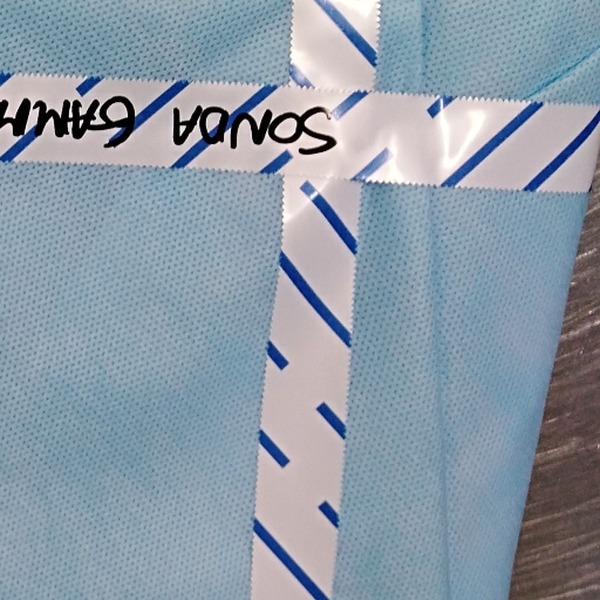}
        &
        \includegraphics[width=4cm,height=4cm]{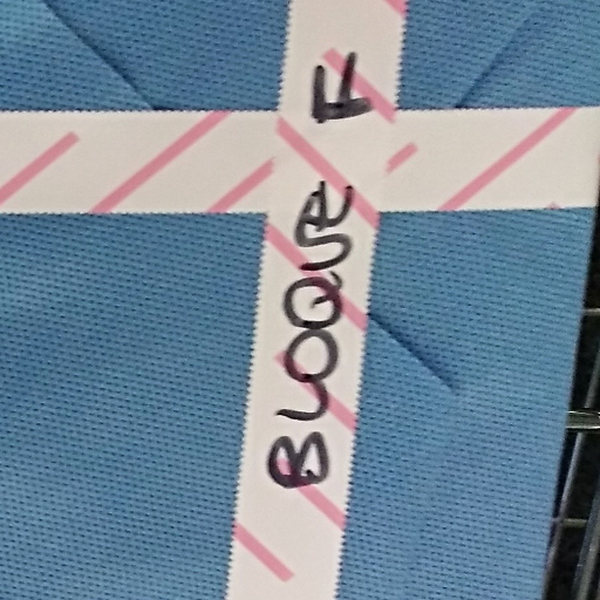}
    \end{tabular}
    \caption{Objects to be detected in the quality-control application case: (top) from left to right, box/bag tracking label, yellowish seal and white-dotted texture related to the presence of a whitish internal filter; (bottom) three kinds of paper tape (black-, blue-, and pink-stripped).}
    \label{fig:targets_qc}
\end{figure}

In the second case, we deal with the detection of a number of control elements that the sterilization unit of a hospital places in boxes and bags containing surgical tools that surgeons and nurses have to be supplied with prior to starting surgery. These elements provide evidence that the tools have been properly submitted to the required cleaning processes, i.e. they have been placed long enough inside the autoclave at a certain temperature, what makes them change their appearance. Figure~\ref{fig:targets_qc} shows, from left to right and top to bottom, examples of six kinds of elements to be detected for this application: the label/bar code used to track a box/bag of tools, the yellowish seal, the three kinds of paper tape which show the black-, blue- and pink-stripped appearance that can be observed in the figure, and an internal filter which is placed inside certain boxes and creates the white-dotted texture that can be noticed (instead of black-dotted when the filter is missing). All these elements, except the label, which is only for box/bag recording and tracking purposes, aim at corroborating the sterilization of the surgery tools contained in the box/bag. Finally, all of them may appear anywhere in the boxes/bags and in a different number, depending on the kind of box/bag.

{Both sorts of applications may require fast operation and are related with target detection: in the quality control case, the aim is to check the correct preparation of a specific product (by detecting the presence of certain control elements) while, in the inspection case, the aim is to recognize anomalous situations for an earlier prevention of undesirable consequences. As can be observed, in our benchmark, the quality control problem involves the detection of man-made, regular objects, though exhibiting very different shapes, sizes and aspect ratios within cluttered scenes, while the visual inspection problem requires the localization of image areas of irregular shape and a large variety of sizes in search of defects. As will be shown, our approach succeeds in both cases, despite the particular challenges and the differences in the objects of interest between them.}

\section{Detector Overview}
\label{sec:overview}

The detector proposed in this work comprises two stages. The first stage is intended to regress unoriented bounding boxes by means of a variant of SSD. As already mentioned, 
SSD is a one-stage object detection approach that makes use of the standard VGG-16 network as backbone though modified by replacing the last fully connected layers and incorporating additional convolutional layers (see~\cite{liu2016ssd} for the details). In comparison with most detection algorithms based on R-CNN, such as~\cite{girshick2015fast} and~\cite{ren2015faster}, SSD does not require any extra procedure to generate proposals. Alternatively, a mechanism of prior boxes is used, from which offsets are regressed for enhanced localization accuracy. On the other side, unlike R-CNN methods, detections are obtained at several scales from a number of layers of the backbone, namely \emph{conv4\_3}, \emph{fc7}, \emph{conv8\_2}, \emph{conv9\_2}, \emph{conv10\_2} and \emph{conv11\_2}. The corresponding feature maps are subsequently involved in the calculation of a multi-task loss function to regress the parameters of the bounding boxes (i.e. offsets relative to the prior boxes shape, as already said), and to obtain confidence values for the classes. To this end, predicted boxes have to be matched with true bounding boxes to train the detector, and only those positives with enough overlap contribute to the loss, while positives and negatives contribute to the classification loss (after a hard negative mining process to keep the positive \textit{vs} negative samples ratio at 1:3).

For the first stage of the detector, in this work, we follow the proposals-free approach of SSD together with the selection of a set of prior boxes, though with a number of differences: (a) the backbone consists in a pyramid of feature maps involving information at more scales than SSD, as depicted in Fig.~\ref{fig:fpssd}, to favour the detection of both large and small targets; (b) the pyramid involves a map fusion scheme that leads to the best performance among a total of four alternatives; and (c) the set of prior boxes are not arbitrarily hand-picked but the selection is guided by the data, resulting from a clustering procedure taking the ground truth as input. The details can be found in Sections~\ref{sec:fpssd} and~\ref{sec:default_boxes}.

The second stage of the detector consists in a specifically designed network trained to regress the parameters of the rotated bounding box maximally contained in the unoriented bounding boxes stemming from the first stage. A detailed description of this stage is given in Section~\ref{sec:rbox_regression}. 

\begin{figure}[t]
  \centering
  \begin{tabular}{cc}
    \includegraphics[height=6cm]{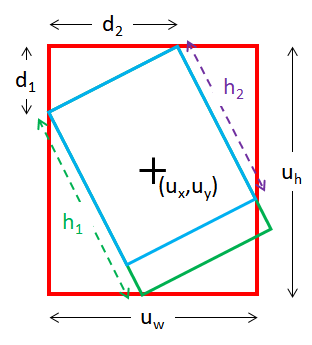} &
    \includegraphics[height=6cm]{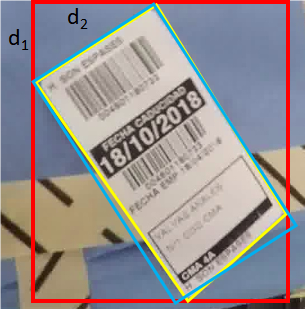}
  \end{tabular}
  \caption{(left) Parameterization of oriented and unoriented bounding boxes: a $(d_1,d_2)$ pair can lead to two different oriented boxes, with heights $h_1$ and $h_2$. (right) Illustration of the ground truth generation process: (yellow) initial 4-side polygon, (red) minimally enclosing unoriented box, (blue) final oriented box.}
  \label{fig:boxes1} 
\end{figure}

To finish, Fig.~\ref{fig:boxes1}(left) illustrates how bounding boxes are parametrized in our approach. For unoriented bounding boxes, we make use of the standard parameters, namely the box center $(u_x,u_y)$ and its width $u_w$ and height $u_h$. Regarding oriented bounding boxes, they are expressed in terms of the unoriented bounding box they are defined in, by means of intercepts $(d_1,d_2)$. As shown in Fig.~\ref{fig:boxes1}(left), these intercepts result from the intersection between the rotated box sides and the unrotated box sides. Since this parametrization can lead to two different rotated boxes, a third optional parameter $h$ can be included in the definition of the oriented bounding box to disambiguate between $h_1$ and $h_2$. 

Given this parametrization, a particular procedure is followed for generating the ground truth necessary for training, which is illustrated in Fig.~\ref{fig:boxes1}(right). First, a four-side polygon is defined over the target (in yellow), from which the minimal unrotated bounding box enclosing is found (in red), to finally define the inside rotated box from intercepts $d_1$ and $d_2$ (in blue).

\section{Feature Pyramid Single Shot Multi-box Detector (FPSSD)}
\label{sec:fpssd}

SSD uses feature maps from different layers of the network to regress bounding boxes. More precisely, SSD adopts large-scale feature maps to detect small targets, and conversely uses small-scale feature maps to detect large targets. In this work, we additionally make use of the \textit{feature pyramid} concept to fuse feature maps from top layers with feature maps from bottom layers to obtain enhanced features containing both semantic information and detailed features, what is exploited to detect different scale targets. 

The idea of the feature pyramid originates from the \textit{image pyramid} concept, which aims at being able to analyze an image at multiple scales by means of multi-scale sampling of the original image via e.g. Gaussian kernels. As assisted by a hierarchical CNN, a \textit{feature pyramid} can be built in one single feed-forward pass that \textit{simultaneously} calculates the multi-scale features of the input image. Hence, the feature pyramid can efficiently address the multi-scale problem with a relative cost. 

\begin{figure}[t]
  \centering
  \begin{tabular}{@{}lll@{}}
  \includegraphics[height=2.4cm]{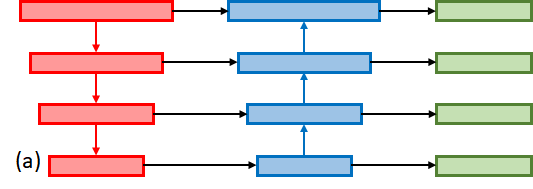} &
  \multicolumn{2}{l}{
  \includegraphics[height=2.4cm]{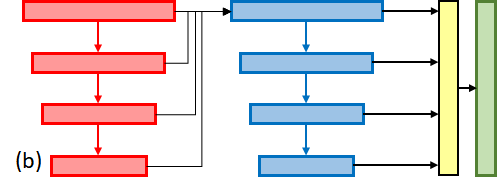}
  } \\
  \includegraphics[height=2.4cm]{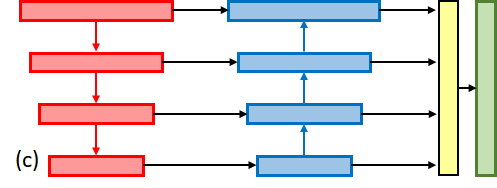} &
  \includegraphics[height=2.4cm]{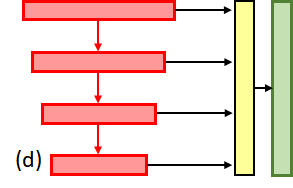} &
  \includegraphics[height=2.4cm]{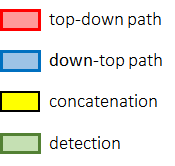}
  \end{tabular}
  \caption{Different strategies for fusing feature maps in a feature pyramid: (a) feature maps are fussed from top to bottom layer by layer; (b) a lightweight architecture that merges feature maps from top to bottom; (c) FPSSD; (d) original SSD approach, which uses feature maps from different layers separately.}
  \label{fig:com_fpn_arch}
\end{figure}

So far, several works have implemented the feature pyramid concept onto DCNNs (see~\cite{lin2017feature,li2018fssd,fu2017dssd}, among others). The four typical approaches for fusing the feature maps are overviewed in Fig.~\ref{fig:com_fpn_arch}. Figure~\ref{fig:com_fpn_arch}(a) illustrates the most common strategy, FPN, which merges feature maps layer by layer by element-wise addition and performs detection from each scale/feature map. Another method is the lightweight fusion strategy named FSSD~\cite{li2018fssd} shown in Fig.~\ref{fig:com_fpn_arch}(b). In this case, features from different layers at different scales are concatenated together first and used next to generate a series of pyramid features. Lastly, the different feature maps are combined by the concatenation layer and sent to the loss function. Though this method is capable of saving computational costs as compared to method (a), the feature maps feeding the detector finally lack certain semantic information. Figure~\ref{fig:com_fpn_arch}(c) illustrates FPSSD, our method, which employs a strategy identical to FPN to fuse the feature maps, but aims at reducing the computational cost by means of a concatenation layer that combines the different feature maps. Subsequently, the combined feature maps are fed into the detector. Lastly, Fig.~\ref{fig:com_fpn_arch}(d) depicts the strategy adopted in the original SSD. Among others, it shows that SSD does not integrate any feature fusion module, and thus it has a limited capability to capture simultaneously low-level details and high-level semantic data. 

\begin{figure}
  \centering
  \begin{tabular}{@{\hspace{0mm}}l|l|l|l@{\hspace{0mm}}}
  \multicolumn{4}{c}{\includegraphics[width=0.96\textwidth, clip=true, trim=0 0 20 0]{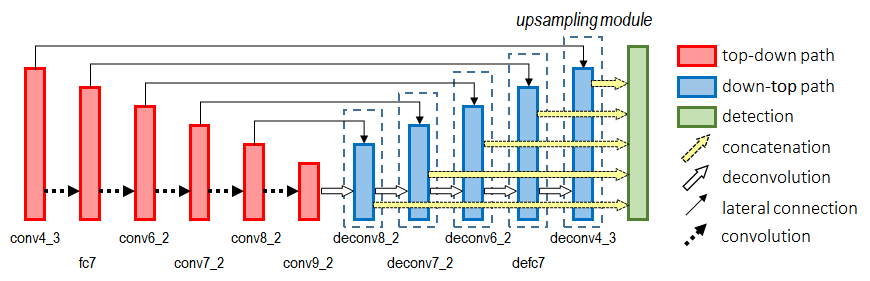}} \\
  \hline 
  \scriptsize (a) & \scriptsize (b) & \scriptsize (c) & \scriptsize (d) \\[-4mm]
  \includegraphics[width=0.22\textwidth]{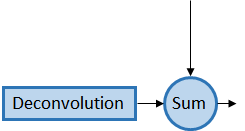} &
  \includegraphics[width=0.22\textwidth]{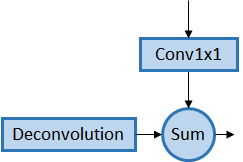} &
  \includegraphics[width=0.22\textwidth]{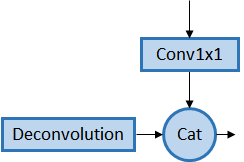} &
  \includegraphics[width=0.22\textwidth]{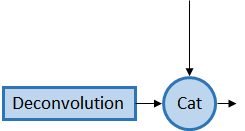}
  \end{tabular}
  \caption{FPSSD: (top) architecture, (bottom) alternative implementations of the upsampling modules (\textsl{Sum}, \textsl{Cat} and \textsl{Conv1$\times$1} respectively denote pixel-wise sum, concatenation and 1$\times$1 convolution).}
  \label{fig:fpssd}
\end{figure}

Figure~\ref{fig:fpssd}(top) outlines the architecture of FPSSD. As can be observed, the feature maps are extracted from the \emph{conv4\_3}, \emph{fc7}, \emph{conv6\_2}, \emph{conv7\_2}, \emph{conv8\_2}, and \emph{conv9\_2} layers of the original SSD network~\cite{Liu2016}. On the other side, deconvolution layers are utilized to enlarge the respective feature maps. We also make use of $1\times 1$ convolutional layers, termed as lateral connections in~\cite{lin2017feature}, to unify the output channels of all feature maps. Lastly, down-top layers integrate different scales and submit the result to the detector to predict the category and localization of targets. 

\section{Selection of (unoriented) Prior Boxes}
\label{sec:default_boxes}

\begin{table}
    \caption{Mean {AIOU} (mIOU) \textit{vs} number of prior boxes, selection method and task. {(See Section~\ref{sec:metrics} for a description of the metrics employed.)}}
    \centering
    \small
    \begin{tabular}[width=0.8\linewidth]{clcc}
          \toprule
          Task & Approach & no. prior boxes & mIOU (\%) \\
          \midrule
          \multirow{7}{*}{Quality Control}
          & Hand-Picked & 4 & 36.75  \\
          & Hand-Picked & 5 & 42.51  \\
          & Hand-Picked & 6 & 49.53  \\
          & Hand-Picked & 10 & 55.44  \\
          & Clustering & 4 & 58.05 \\
          & Clustering & 5 & 61.70 \\
          & Clustering & 6 & \bf 63.56 \\
          \midrule
          \multirow{7}{*}{Inspection}
          & Hand-Picked & 4 & 35.93  \\
          & Hand-Picked & 5 & 37.96  \\
          & Hand-Picked & 6 & 42.75  \\
          & Hand-Picked & 10 & 61.82  \\
          & Clustering & 4 & 61.58 \\
          & Clustering & 5 & 63.37 \\
          & Clustering & 6 & \bf 65.31 \\
          \bottomrule
    \end{tabular}
    \label{tab:cluster}
\end{table}

SSD predefines a total of 6 prior boxes per feature map location by imposing different size combinations $(w_k,h_k)$ manually picked. Since, on the one hand, the shape of the bounding boxes to detect can vary significantly and, on the other hand, SSD regresses the predicted bounding boxes from the prior boxes, a proper selection of those prior boxes becomes crucial for achieving a high detection success; as already noted in~\cite{Redmon2017}, such a proper selection contributes to the stability of the underlying optimization process, converges faster and improves effectively the \textit{Intersection over Union} (IOU) between predicted and true boxes. Hence, our object detector makes use of prior boxes selected automatically in accordance to the available data.

In more detail, we run the well-known K-means algorithm over the bounding boxes belonging to the ground truth, using box width and height as the clustering features. Instead of the Euclidean distance, typically used by K-means implementations, we define IOU as a distance metric because we have observed better clustering results with the latter. The distance between a sample box $b_i$ and the cluster centroid $c_j$ is hence defined as: 
\begin{align}
d(b_i,c_j) 
&= 1 - \text{IOU}(b_i,c_j) 
= 1 - \frac{b_i \cap c_j}{b_i \cup c_j}
= 1 - \frac{o(b_i,c_j)}{a(b_i) + a(c_j) - o(b_i,c_j)}
\label{eq:cluster}
\end{align}
where $o(\cdot,\cdot)$ denotes overlapping area and $a(\cdot)$ denotes area. 

Table~\ref{tab:cluster} shows averages of the IOU metric (see Section~\ref{sec:metrics}) for hand-picked prior boxes and automatically selected boxes by clustering, for both detection tasks, inspection and quality control, and a different number of prior boxes (for the hand-picked cases, we predefine the boxes similarly to SSD). We can see that 4 clusters automatically selected yield better performance than 10 hand-picked prior boxes. This means that it is possible to propose automatically higher-quality and better parameterized prior boxes. As could be expected, the more clusters, the better is the performance (the trend can be observed to continue for 7 or more prior boxes), although the number of clusters should not be high to keep reasonable the running time.

\section{Regression of Oriented Bounding Boxes}
\label{sec:rbox_regression}

\begin{figure}
  \centering
  \includegraphics[width=0.9\columnwidth]{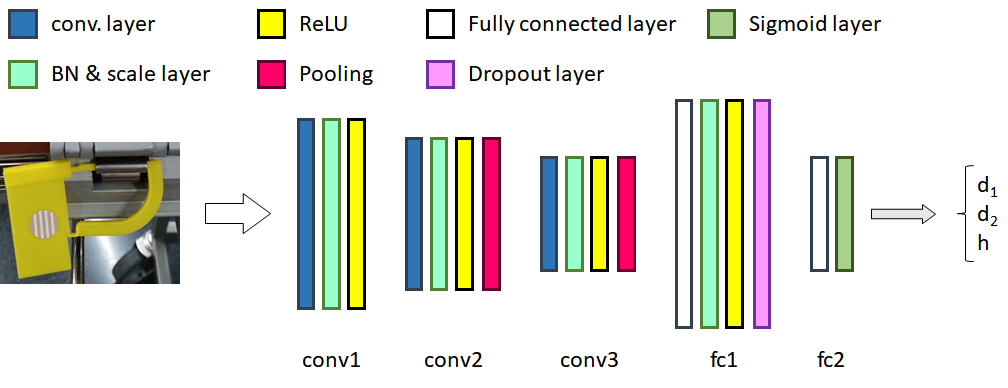}
  \caption{Architecture of the RBox regression network.}
  \label{fig:rbox_regress_network}
\end{figure}

To regress the parameters of the rotated boxes, a lightweight convolutional network based on LeNet~\cite{LeCun1998} has been adopted. With regard to the original network, the rotated boxes (RBox) regression network exhibits several differences: (1) the input size is $63\times63$ after the incorporation of an additional convolutional layer at the beginning of the network, in order to avoid reducing the image to LeNet's $28\times28$ input pixels and lose information; (2) batch normalization is used after each convolutional layer to speed up convergence during training (this has also been shown to decrease the effect of covariate shift from the hidden layers~\cite{Ioffe2015}); (3) since the bounding boxes parameters $(d_1,d_2,h)$ range from 0 to 1, a sigmoid layer lies between the last fully connected layer and the loss layer; and (4) lastly, an Euclidean distance loss layer is used during regression:
\begin{equation}
  L(d,g) = \frac{1}{2N}\sum_{i=1}^N (d_1^i - g_{d_1}^i)^2 + (d_2^i - g_{d_2}^i)^2 + (h^i - g_h^i)^2
  \label{eq:rboxloss}
\end{equation}
where $d = (d_1, d_2, h)$ denotes the predicted offsets and height, $g = (g_{d_1}, g_{d_2}, g_{h})$ represents the ground truth and $N$ is the size of the mini-batch. The architecture of the RBox regression network can be found in Fig.~\ref{fig:rbox_regress_network}. 

\section{Experimental Results and Discussion}
\label{sc:results}

\subsection{Experimental Setup}

The FPSSD and RBox networks have been implemented using Caffe~\cite{jia2014caffe}. Referring particularly to FPSSD, the VGG-16 network is taken as the backbone, indentically to the original SSD. 

Trainings and all experiments have been performed on a PC platform fitted with an Intel i9-9900K processor with 64Gb RAM and an Nvidia RTX 2080Ti GPU. The quality control dataset consists of 484 images for a total of 7 categories, while the inspection dataset comprises 214 images for corrosion detection. All the images have been resized to $512\times 512$ pixels.  

As for training, we have adopted a multiple steps strategy, where the learning rate was set to $10^{-5}$ during the first 8000 iterations, the next 6000 iterations used a learning rate of $10^{-6}$, and the final 6000 iterations employed a learning rate of $10^{-7}$. The batch size was set to 10, which is the best configuration for the GPU involved in the experiments. We have employed SGD for network optimization, and the weight decay and the momentum were set to 0.001 and 0.9, respectively. As a compromise between accuracy and computation time, object detection was performed using six prior boxes whose features resulted from the clustering process described in Section~\ref{sec:default_boxes}.

\subsection{Assessment Metrics}
\label{sec:metrics}

We employ the following metrics for performance evaluation:
\begin{itemize}
\item {For both unoriented and oriented bounding boxes, we consider the Recall (R), the Precision (P) and the Average Precision (AP), the latter measured as the area under the P-R curve for a set of recall values~\cite{Everingham2010}. Unless otherwise stated, detected bounding boxes with a confidence above 0.7 have been considered as the predictions of the detector (as usual for object detection).} 
\item {For the multi-class case, we consider both the value of a metric for a specific class as well as the average of the respective values for each class, i.e. the mean Recall, the mean Precision, the corresponding F$_1$ score and the mean AP, all calculated by means of:
\begin{equation}
\text{m}\mathcal{M} = \frac{1}{C} \sum_{c=1}^C \mathcal{M}_c,
\end{equation} 
where $\mathcal{M}_c$ is the value of metric $\mathcal{M}$ for class $c$, and $C$ is the number of classes.}
\item {For the unoriented bounding boxes, we also consider the mean IOU as the average over all categories of the averaged IOU (AIOU): 
\begin{equation}
\text{mIOU} = \frac{1}{C} \sum_{c=1}^C \text{AIOU}_c,\ \text{with } \text{AIOU}_c = \frac{\displaystyle\sum_{b_j \in c} \text{IOU}(b_j,g_j)}{n_c} = \frac{1}{n_c}\sum_{b_j \in c} \frac{b_j\cap g_j}{b_j \cup g_j}
\label{eq:mIOU}
\end{equation}
where $b_j \in  c$ is the set of predictions $b_j$ for class $c$, whose cardinality is $n_c$, and $g_j$ is the true bounding box with highest overlap with $b_j$.}
\item To determine the performance of oriented detection, we also provide the averaged RBox IOU (ARIOU) and the mean {ARIOU} (mRIOU) as supplementary performance metrics (analogously to Eq.~\ref{eq:mIOU}). Unlike the case of unoriented bounding boxes, the shape of the intersection of two rotated bounding boxes turns out to be into a convex polygon. In general, the area $A_\text{cp}$ of such a polygon is given by:
\begin{equation}
A_\text{cp} = \frac{1}{2} \sum_{i=1}^n (x_{i}y_{i+1} - x_{i+1}y_i) \quad \text{[Shoelace Formula]}
\end{equation}
where $\{(x_1,y_1),\dots,(x_n,y_n)\}\}$ are the coordinates of the polygon vertices arranged counterclockwise, and $(x_{n+1},y_{n+1}) = (x_1,y_1)$.
\item {To finish, in order to measure the accuracy of the parameters regressed, we also adopt the mean absolute error (MAE) for the regression targets considered, calculated as follows:
\begin{equation}
\text{MAE} = \frac{1}{N}\sum\limits_{i=1}^{N} \sum\limits_{j} |t_{p,j}^{(i)} - t_{g,j}^{(i)}|
\end{equation}
where $t_{p,j}^{(i)}$ and $t_{g,j}^{(i)}$ respectively denote the $i$-th prediction for target $j$ and the corresponding ground truth, and $N$ is the number of predictions.}
\end{itemize}

\subsection{Regression Results for Unoriented Bounding Boxes}
\label{sec:fpssd_bbox_detection_results}

In this section, we report on the performance obtained for unoriented bounding boxes detection. We start with an ablation study considering the effect of the lateral connections between layers of the top-down and dowm-top paths and the necessary map fusion approaches. We consider SSD 512 as a baseline and the alternatives that are enumerated in Fig.~\ref{fig:fpssd}(bottom), which contemplate pixel-wise sum and concatenation for map fusion, and the use or not of 1 $\times$ 1 convolutional filters to unify the number of output channels from top to bottom layers (different for the VGG-16 network). Results for different metrics are reported in Table~\ref{tab:fpssd_ablation}. As can be observed, for both tasks, option (b) attains the largest performance in almost all cases, except for the quality control task and the mean precision, although the second best is also for configuration (b) at a very short distance.

\begin{table}[t]
    \caption{Ablation study: effect of lateral connections and the feature map fusion approach. (Bold face denotes best.)}
    \centering
    \footnotesize
    \begin{tabular}[width=1.0\linewidth]{@{\hspace{0mm}}l@{\hspace{1mm}}ll@{\hspace{2mm}}c@{\hspace{2mm}}c@{\hspace{2mm}}c@{\hspace{2mm}}c@{\hspace{0mm}}}
        \toprule
        Task & Configuration & Fig. & R & P & F$_1$ & AP  \\
        \midrule
        \multirow{5}{1.8cm}[-2mm]{Visual Inspection}
        & SSD 512                           &                    &     0.8311 &     0.9434 &     0.8837 &     0.8218\\
        & FPSSD 512 + Sum                   & \ref{fig:fpssd}(a) &     0.8241 &     0.9513 &     0.8831 &     0.8131\\
        & FPSSD 512 + 1$\times$1 conv + Sum & \ref{fig:fpssd}(b) & \bf 0.9113 & \bf 1.0000 & \bf 0.9536 & \bf 0.9091\\
        & FPSSD 512 + 1$\times$1 conv + Cat & \ref{fig:fpssd}(c) &     0.8264 &     0.9433 &     0.8810 &     0.8133\\
        & FPSSD 512 + Cat                   & \ref{fig:fpssd}(d) &     0.8262 &     0.9563 &     0.8865 &     0.8172\\
        \midrule
        \multirow{5}{1.8cm}[-2mm]{Quality Control}
        & SSD 512                           &                    &     0.8111 &     0.9663 &     0.8819 &     0.8055\\
        & FPSSD 512 + Sum                   & \ref{fig:fpssd}(a) &     0.8543 &     0.9324 &     0.8916 &     0.8471\\
        & FPSSD 512 + 1$\times$1 conv + Sum & \ref{fig:fpssd}(b) & \bf 0.8715 &     0.9550 & \bf 0.9113 & \bf 0.8644\\
        & FPSSD 512 + 1$\times$1 conv + Cat & \ref{fig:fpssd}(c) &     0.8682 & \bf 0.9579 &     0.9108 &     0.8632\\
        & FPSSD 512 + Cat                   & \ref{fig:fpssd}(d) &     0.8646 &     0.9124 &     0.8879 &     0.8583\\
        \bottomrule
    \end{tabular}
    \label{tab:fpssd_ablation}
\end{table}

\begin{table}[t]
    \caption{Performance results of FPSSD and SSD for both tasks. (Bold face denotes best.)}
    \centering
    \small
    \begin{tabular}[width=.2\columnwidth]{@{\hspace{0mm}}l@{\hspace{2mm}}l@{\hspace{2mm}}ccccc@{\hspace{0mm}}}
        \toprule
        Task                             & Class          & R          & P           & F$_1$      &  AP        & AIOU   \\
        \midrule                                                               
        Visual Inspection \\(FPSSD 512)  & Corrosion      & \bf 0.9113 & \bf 1.0000  & \bf 0.9536 & \bf 0.9091 & \bf 0.9375 \\
        \midrule                                                                      
        Visual Inspection \\(SSD 512)    & Corrosion      &     0.8311 &     0.9434  &     0.8837 &     0.8218 &     0.8486 \\
        \midrule                                                                   
        \multirow{4}{2.5cm}{Quality\\ Control\\ (FPSSD 512)}                            
                                        & Label           & \bf 0.9177 & \bf 0.9779  & \bf 0.9468 & \bf 0.9097 & \bf 0.8707 \\
                                        & Seal            & \bf 0.8566 &     0.9697  & \bf 0.9096 & \bf 0.8461 & \bf 0.8382 \\
                                        & Black tape      & \bf 0.7191 &     0.8793  & \bf 0.7912 & \bf 0.7055 & \bf 0.7695 \\
                                        & Blue tape       & \bf 0.9139 & \bf 0.9421  & \bf 0.9278 & \bf 0.9093 & \bf 0.8468 \\
                                        & Pink tape       & \bf 0.8219 &     0.9614  & \bf 0.8862 & \bf 0.8206 &     0.8236 \\
                                        & Internal filter & \bf 1.0000 & \bf 1.0000  & \bf 1.0000 & \bf 1.0000 & \bf 0.9166 \\
        \midrule                                                                      
                                        & Mean value      & \bf 0.8715 &     0.9550  & \bf 0.9113 & \bf 0.8644 & \bf 0.8443 \\
        \midrule                                                                      
        \multirow{4}{2.5cm}{Quality\\ Control\\ (SSD 512)}                             
                                        & Label           &     0.8821 &     0.9691  &     0.9236 &     0.8783 &     0.8484 \\
                                        & Seal            &     0.8301 & \bf 0.9769  &     0.8975 &     0.8289 &     0.8256 \\
                                        & Black tape      &     0.5468 & \bf 0.9342  &     0.6898 &     0.5387 &     0.7673 \\
                                        & Blue tape       &     0.8839 &     0.9328  &     0.9077 &     0.8773 &     0.8346 \\
                                        & Pink tape       &     0.7387 & \bf 0.9668  &     0.8375 &     0.7261 & \bf 0.8259 \\
                                        & Internal filter &     0.9841 & \bf 1.0000  &     0.9920 &     0.9841 &     0.9111 \\
        \midrule                                                                      
                                        & Mean value      &     0.8111 & \bf 0.9663  &     0.8819 &     0.8055 &     0.8404 \\
        \bottomrule
    \end{tabular}
    \label{tab:fpssd_performance}
\end{table}

Table~\ref{tab:fpssd_performance} breaks down into the different categories the performance values of Table~\ref{tab:fpssd_ablation} for configuration (b). As can be observed, for the quality control task, FPSSD yields better performance for all objects and almost all metrics, including the intersection over union. SSD is slightly better only regarding precision; {the F$_1$ score, which combines P and R, is also higher for FPSSD in all cases}. Major differences correspond to the different kinds of paper tape, with differences highly noticeable for recall, the F$_1$ score and the AP. As for the visual inspection task, the differences are even larger and take place for all metrics.

\begin{figure*}
    \begin{center}
    \begin{tabular}{@{\hspace{0mm}}c@{\hspace{1mm}}c@{\hspace{1mm}}c@{\hspace{1mm}}c@{\hspace{1mm}}c@{\hspace{0mm}}}
        \rotatebox{90}{FPSSD} 
        &
        \includegraphics[width=0.23\columnwidth,height=0.23\columnwidth]{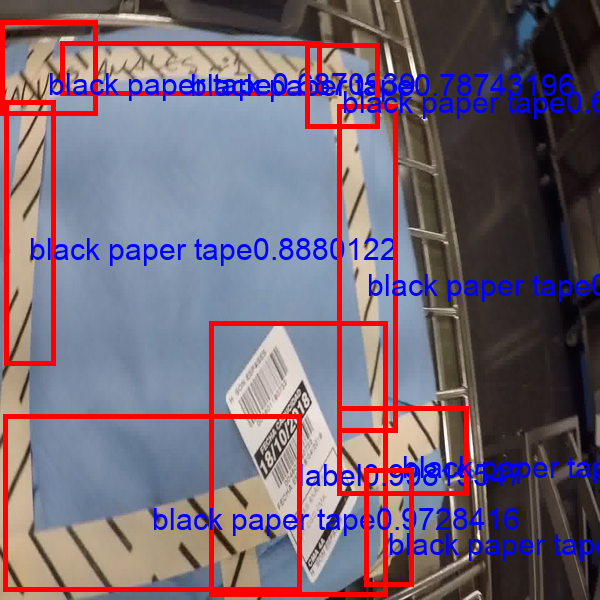}
        &
        \includegraphics[width=0.23\columnwidth,height=0.23\columnwidth]{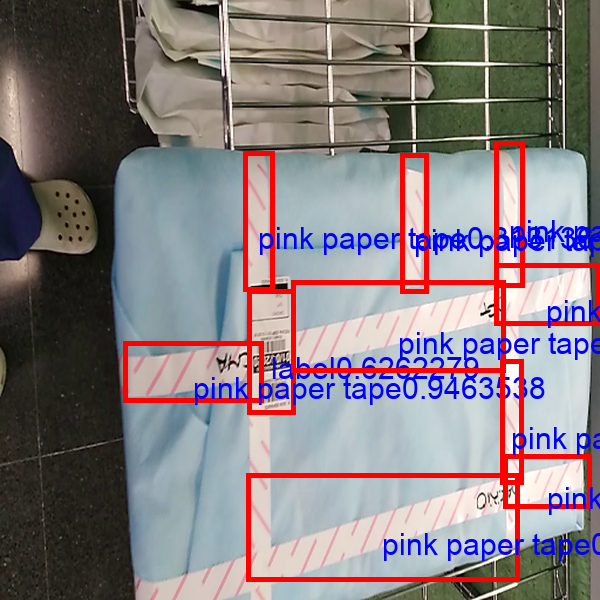}
        &
        \includegraphics[width=0.23\columnwidth,height=0.23\columnwidth]{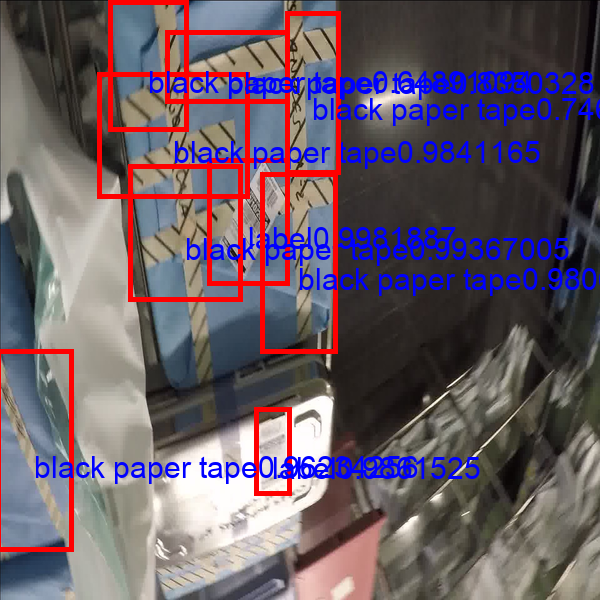}
        &
        \includegraphics[width=0.23\columnwidth,height=0.23\columnwidth]{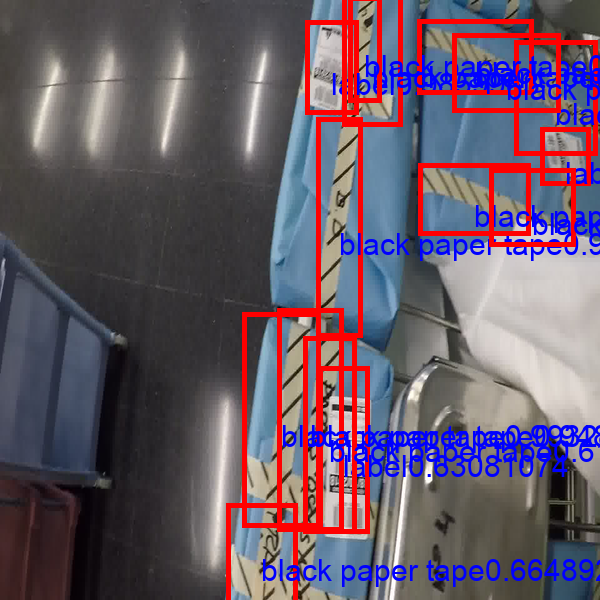}
        \\
        \rotatebox{90}{SSD} 
        & 
        \includegraphics[width=0.23\columnwidth,height=0.23\columnwidth]{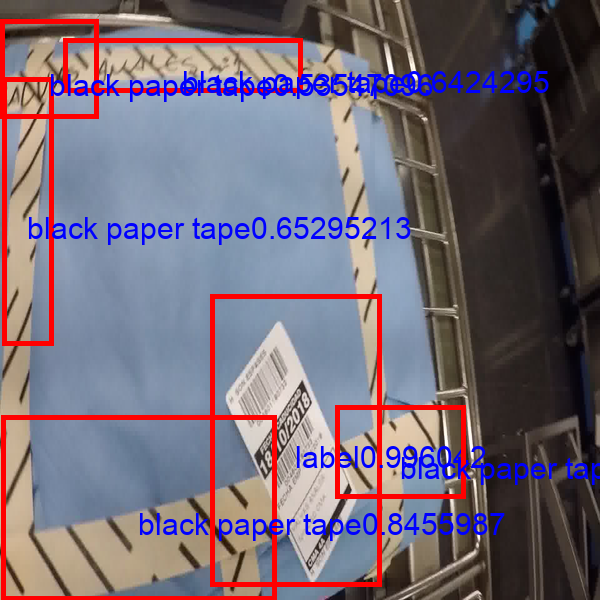}
        &
        \includegraphics[width=0.23\columnwidth,height=0.23\columnwidth]{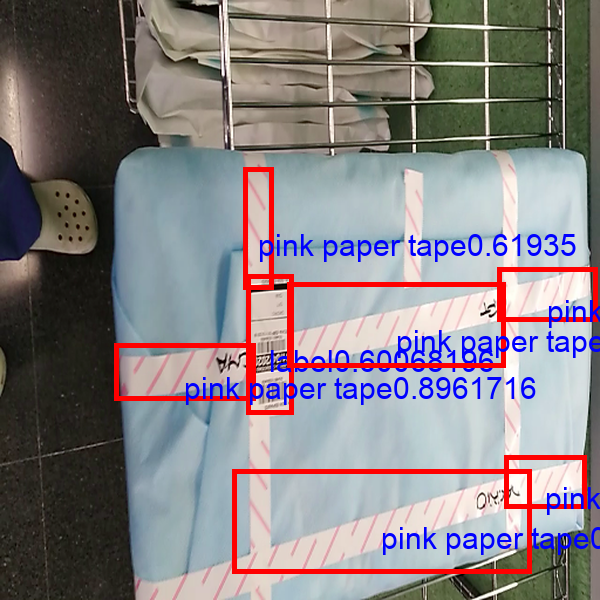}
        &
        \includegraphics[width=0.23\columnwidth,height=0.23\columnwidth]{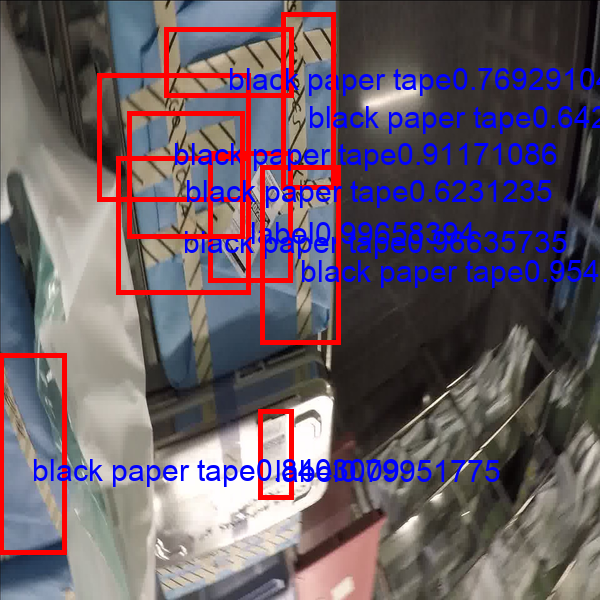}
        &
        \includegraphics[width=0.23\columnwidth,height=0.23\columnwidth]{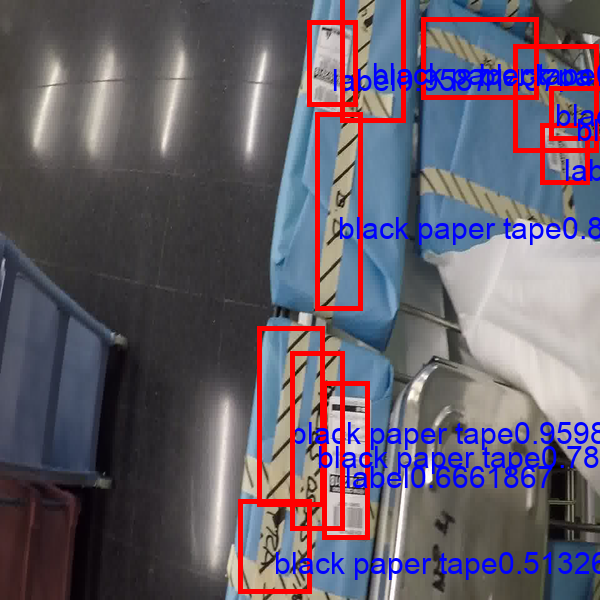}
        \\
        \rotatebox{90}{FPSSD} 
        &
        \includegraphics[width=0.23\columnwidth,height=0.23\columnwidth]{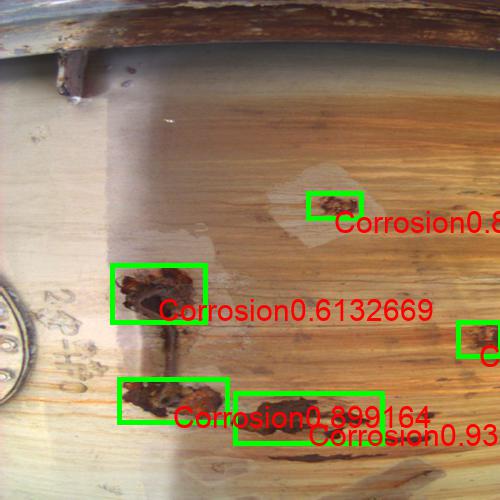}
        &
        \includegraphics[width=0.23\columnwidth,height=0.23\columnwidth]{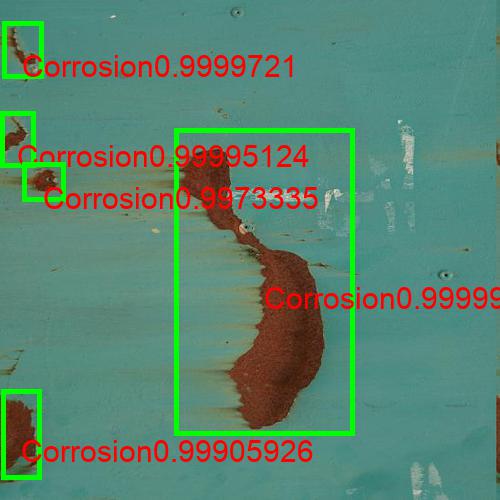}
        &
        \includegraphics[width=0.23\columnwidth,height=0.23\columnwidth]{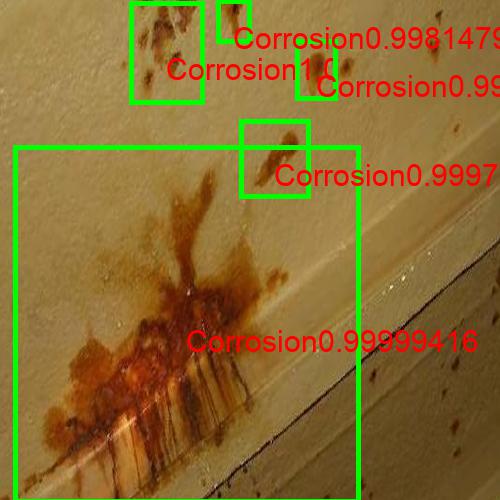}
        &
        \includegraphics[width=0.23\columnwidth,height=0.23\columnwidth]{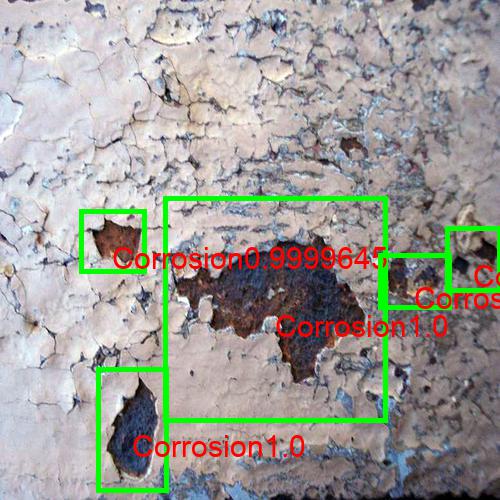}
        \\
        \rotatebox{90}{SSD} 
        &
        \includegraphics[width=0.23\columnwidth,height=0.23\columnwidth]{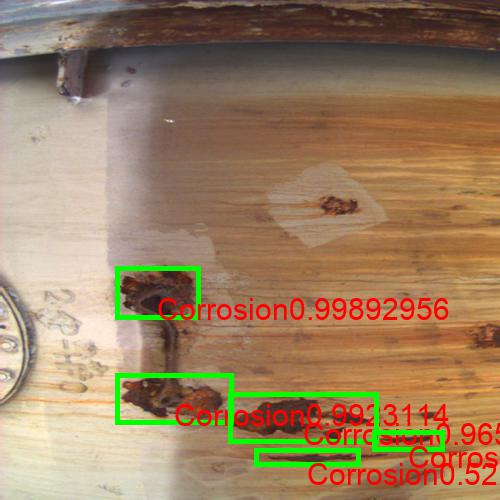}
        &
        \includegraphics[width=0.23\columnwidth,height=0.23\columnwidth]{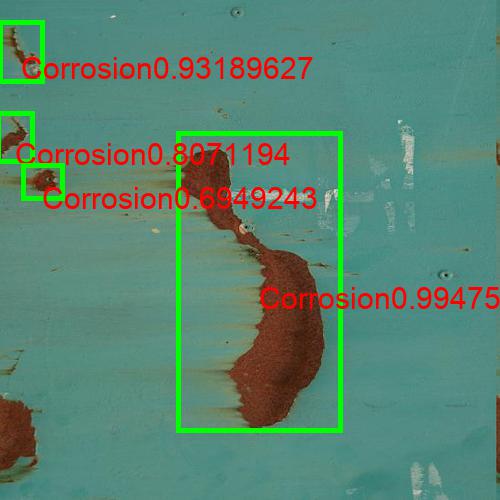}
        &
        \includegraphics[width=0.23\columnwidth,height=0.23\columnwidth]{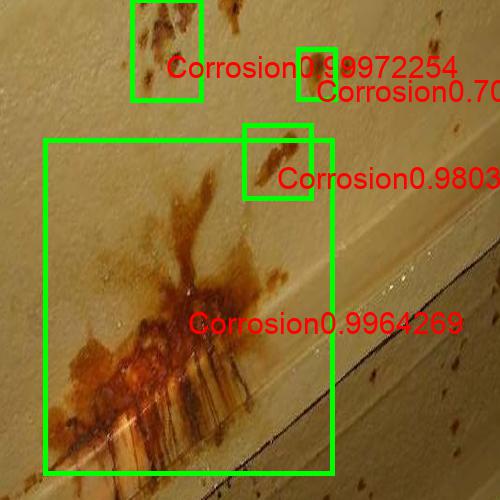}
        &
        \includegraphics[width=0.23\columnwidth,height=0.23\columnwidth]{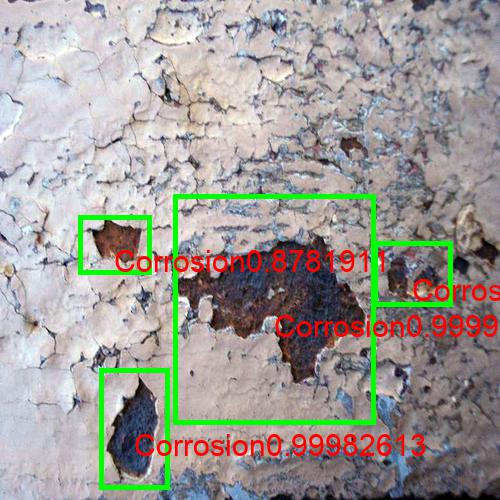}
    \end{tabular}
    \end{center}
    \vspace{-5mm}
    \caption{Unoriented detection results for FPSSD and SSD: (1st \& 2nd rows) FPSSD and SSD for the Quality Control task; (3rd \& 4th rows) FPSSD and SSD for the Visual Inspection task.}
    \label{fig:bbox_results}
\end{figure*}

To finish, some qualitative results from a selection of images from both datasets can be found in Fig.~\ref{fig:bbox_results}. It can be noticed that, even in a very cluttered scene, FPSSD can detect almost all control elements in the quality control task, including the different sorts of paper tape. For the visual inspection task, FPSSD also achieves better results, being able to detect small areas affected by corrosion. FPSSD leads thus to a quite competitive performance on the two tasks.

\subsection{Regression Results for Oriented Bounding Boxes (RBox alone)}
\label{sec:rbox_regression_results}

As already mentioned, though FPSSD leads to good performance for unoriented detection for both tasks, for some elongated targets, either regularly-shaped or irregularly-shaped, the results of FPSSD can be inaccurate, apart from the fact that unoriented bounding boxes tend to include parts of other objects and even cover a large fraction of the {background} in order to fully contain certain objects of interest (this is specially true for the quality control case, but can also happen for the visual inspection case, as shown in Fig.~\ref{fig:bbox_results}). {These} are the reasons why oriented bounding boxes are considered in this work. In this section, we analyze the performance of the RBox regression network described in Section~\ref{sec:rbox_regression}.

For a start, Table~\ref{tab:ablation_rbox} shows the MAE for each regression target and two configurations: (a) two-target regression $(d_1,d_2)$ and (b) three-target regresssion $(d_1,d_2,h)$. As can be observed, the MAE values for $d_1$ and $d_2$ for the two-target case are lower than the corresponding MAE values for the three-target case. Moreover, the average MAE of the two-target case is also lower than the average MAE for the three-target case. {On the other side, we have trained AlexNet~\cite{krizhevsky2017imagenet} for the two tasks after replacing the final softmax layer by a sigmoid layer. The resulting model has been used as a baseline to compare with. As Table~\ref{tab:ablation_rbox} also shows, AlexNet produces, on average, worse predictions than the RBox network.} 


\begin{table}
    \caption{MAE values for the regression targets considered by the RBox network in comparison with AlexNet. (Bold face denotes best.)}
    \centering
    \small
    \begin{tabular}[width=.3\columnwidth]{llcccc}  
        \toprule
        Task & Approach             & $d_1$     & $d_2$     & $h$      & average \\
        \midrule
        \multirow{2}{*}[-4.5mm]{Quality Control}
                & RBox (2-target)     & \bf 0.1059 & \bf 0.1017 & -          & \bf 0.1038 \\
                & RBox (3-target)     &     0.2289 &     0.1862 & \bf 0.0557 &     0.1569 \\
                & AlexNet (2-target)  &     0.2038 &     0.1915 & -          &     0.1976 \\
                & AlexNet (3-target)  &     0.2430 &     0.1997 &     0.0932 &     0.1786 \\
        \midrule
        \multirow{2}{*}[-4.5mm]{Visual Inspection}
                & RBox (2-target)     & \bf 0.1556 & \bf 0.1612 & -          & \bf 0.1584 \\
                & RBox (3-target)     &     0.3151 &     0.3105 & \bf 0.0889 &     0.2381 \\
                & AlexNet (2-target)  &     0.1722 &     0.1915 & -          &     0.1818 \\
                & AlexNet (3-target)  &     0.2722 &     0.3744 &     0.2501 &     0.2989 \\
        \bottomrule
    \end{tabular}
    \label{tab:ablation_rbox}
\end{table}

\begin{figure*}
    \centering
    \begin{tabular}{c@{\hspace{3mm}}c@{\hspace{2mm}}c@{\hspace{1mm}}c}
        \footnotesize (a) RBox, 2-tar.
        &
        \footnotesize (b) RBox, 3-tar.
        &
        \footnotesize (c) AlexNet, 2-tar.
        &
        \footnotesize (d) AlexNet, 3-tar.
        \\
        \includegraphics[width=0.20\columnwidth,height=0.20\columnwidth]{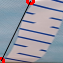}
        &
        \includegraphics[width=0.20\columnwidth,height=0.20\columnwidth]{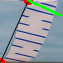}
        &
        \includegraphics[width=0.20\columnwidth,height=0.20\columnwidth]{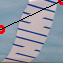}
        &
        \includegraphics[width=0.20\columnwidth,height=0.20\columnwidth]{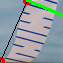}
        \\
        \includegraphics[width=0.20\columnwidth,height=0.20\columnwidth]{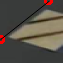}
        &
        \includegraphics[width=0.20\columnwidth,height=0.20\columnwidth]{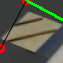}
        &
        \includegraphics[width=0.20\columnwidth,height=0.20\columnwidth]{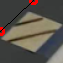}
        &
        \includegraphics[width=0.20\columnwidth,height=0.20\columnwidth]{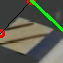}
        \\
        \includegraphics[width=0.20\columnwidth,height=0.20\columnwidth]{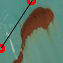}
        &
        \includegraphics[width=0.20\columnwidth,height=0.20\columnwidth]{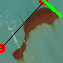}
        &
        \includegraphics[width=0.20\columnwidth,height=0.20\columnwidth]{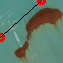}
        &
        \includegraphics[width=0.20\columnwidth,height=0.20\columnwidth]{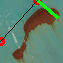}
        \\
        \includegraphics[width=0.20\columnwidth,height=0.20\columnwidth]{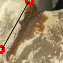} 
        &
        \includegraphics[width=0.20\columnwidth,height=0.20\columnwidth]{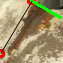} 
        &
        \includegraphics[width=0.20\columnwidth,height=0.20\columnwidth]{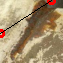} 
        &
        \includegraphics[width=0.20\columnwidth,height=0.20\columnwidth]{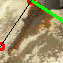}
    \end{tabular}
    \vspace{-2mm}
    \caption{RBox regression results: (a) RBox network for 2 regression targets, (b) RBox network for 3 regression targets, (c) AlexNet for 2 regression targets, (d) AlexNet for 3 regression targets. (The red dots correspond to regression targets $d_1$ and $d_2$, while the green line represents the regression target $h$.)}
    \label{fig:rbox_regression_results}
\end{figure*}

Figure~\ref{fig:rbox_regression_results} shows some examples of detections of rotated bounding boxes for the two tasks, and for two- and three-target regression. In the pictures, the red points correspond to the $d_1$ and $d_2$ intercepts, while the green line represents the third regression target $h$. The black line just connects the red points to show the predicted orientation of the object detected. It can be observed that the black lines in the first column (using two-target regression) adhere better to the orientation of the objects than the detections of the second column (using three-target regression); the two-target RBox network outperforms as well the fine-tuned version of AlexNet. 

\subsection{{Regression Results for Oriented Bounding Boxes (end-to-end prediction)}}
\label{sec:rbox_regression_results}

{After the previous evaluation of the RBox network alone, we have connected the FPSSD and the RBox networks to infer oriented detections end to end, i.e. the input of the RBox regression network are the predictions produced by FPSSD above a given confidence level $\tau_c$. Notice that, because the output of FPSSD is a prediction, it could be slightly displaced with regard to the true object location (which is what has been used for training), increasing hence the challenge of estimating correctly the object orientation.} 

{A first performance evaluation of the full FPSSD-RBox detector is shown in Table~\ref{tab:IoU_rbox}, which compares, by means of ARIOU and mRIOU values, the two-target and three-target RBox regression networks for the two application cases. As already mentioned, the two-target regression variant of RBox gives rise to two predictions (see Fig.~\ref{fig:boxes1}). For this and the next experiments, we always select the largest oriented box. As can be noticed, for all the categories of the quality control task, the ARIOU for the two-target regression variant is higher than the ARIOU of the three-target regression variant, as well as the mRIOU. Referring to the elongated objects (i.e. the three types of paper tape), the resulting ARIOU values for the two-target case are 0.6247, 0.5604, and 0.4993, which are significantly higher than those for the alternative. As for the visual inspection task, the ARIOU for the two-target regression case also turns out to be of highest performance, though the difference is not so notorious. The superiority of the simpler network, already observed from the results reported in Table~\ref{tab:ablation_rbox}, is thus confirmed.} 

\begin{table}
{
    \caption{ARIOU and mRIOU values for the combination of FPSSD and the RBox regression network (FPRB), 2- and 3-target regression networks. (Bold face denotes best.)}
    \centering
    \footnotesize
    \begin{tabular}{@{\hspace{0mm}}l@{\hspace{2mm}}c@{\hspace{2mm}}c@{\hspace{2mm}}c@{\hspace{2mm}}c@{\hspace{2mm}}c@{\hspace{2mm}}c@{\hspace{2mm}}c@{\hspace{2mm}}c@{\hspace{0mm}}}
        \toprule
        \multirow{2}{*}{Task $\rightarrow$}
        & \multicolumn{6}{c}{\multirow{2}{*}{Quality Control}}   & & Visual \\
        & \multicolumn{6}{c}{                                }   & & Insp. \\
        \midrule
        Category $\rightarrow$ &              &                  & Blue             & Pink              & Black             & Intl.             & Mean   &                  \\
        Method $\downarrow$    & Label        & Seal             & Tape             & Tape              & Tape              & Filter            & value  & Corrosion        \\
        \midrule
        FPRB (2-target)   & \textbf{0.7102}   & \textbf{0.7123}  & \textbf{0.6247}  & \textbf{0.5604}   & \textbf{0.4993}   & \textbf{0.7669}   & \bf 0.6456 & \bf 0.5932 \\
        FPRB (3-target)   & 0.5160            & 0.4790           & 0.3136           & 0.2962            & 0.3188            & 0.5917            & 0.4192 & 0.5419          \\
        \bottomrule
    \end{tabular}
    \label{tab:IoU_rbox}
}
\end{table}


{Next, we compare the FPSSD-RBox detector with other orientation-aware detectors: 
	\begin{itemize}
	\item TextBoxes++~\cite{liao2018textboxes++}, a text detector also based on SSD; 
	\item DAL~\cite{Ming2021}, an anchor-based solution that focus on label assignment for enhanced detection and, in this respect, proposes an alternative to IoU to select anchor localization in a better way; and 
	\item Rotated RCNN~\cite{Xie2021}, a two-stage detector that intends to be simpler and hence aims at lower inference times with state-of-the-art detection accuracy.
	\end{itemize} 
}

{For a start, Table~\ref{tab:comp_insp} compares the aforementioned algorithms for the visual inspection task, considering different values for the confidence threshold $\tau_c$. True positives (TP), false positives (FP) and false negatives (FN) are reported (TP + FN = nGT, where nGT are the true detections according to the ground truth and is given in Table~\ref{tab:true_det} for each class), as well as the corresponding precision, recall and F$_1$ values, the AP and the ARIOU. As can be observed, the combination FPSSD-RBox outperforms the alternative algorithms almost in all cases, being the second best when it is not the best performing solution.}

{Tables~\ref{tab:comp_qual_05}\,-\,\ref{tab:comp_qual_08} show the same kind of data for the same values of $\tau_c$ as before but for the quality control task, on a \textit{per class} basis and also averaged over all classes. Table~\ref{tab:comp_qual_07} shows the combination FPSSD-RBox again outperforming for all metrics the other detectors for $\tau_c = 0.7$. For other threshold values, the winner fluctuates among the methods involved in the comparison depending on the performance metric considered.}

{As a measure of success when dealing with the visual inspection and the quality control tasks, Table~\ref{tab:best1} summarizes the previous results highlighting the best and second-best algorithm for each category and on average, as well as for the different metrics considered. For scoring purposes, Table~\ref{tab:best2} counts the number of times each algorithm is the best and second-best in each case. As can be observed, FPSSD-RBox is the algorithm that attains the highest scores the highest number of times for both tasks and considering all classes; Oriented RCNN is next in the ranking, followed by DAL and finally TextBoxes++.}

{On the other side, Table~\ref{tab:times} reports on the inference times per image for each algorithm and the two tasks, showing that FPSSD-RBox turns out to be the fastest (in particular, due to the simplicity of the rotated bounding box regression approach), being able to attain more than 55 FPS, followed by TextBoxes++ (almost 42 FPS), DAL (30 FPS) and, finally, Oriented RCNN as the slowest (15 FPS).}

\begin{table}
	{
		\caption{Number of true detections according to the ground truth (nGT) for all classes of the Visual Inspection and the Quality Control tasks: \textit{corrosion} (COR), \textit{label} (LAB), \textit{blue} (BLU), \textit{pink} (PIN) and \textit{black paper tapes} (BLA), \textit{seal} (SEA) and \textit{internal filter} (IFI).}
		\footnotesize
		\begin{center}
			\begin{tabular}{lccccccc}
				\toprule
				    & COR & LAB & BLU & PIN & BLA & SEA & IFI \\
				\midrule
				nGT & 722 & 390 & 598 & 267 & 541 & 153 & 189 \\
				\bottomrule 
			\end{tabular}
		\end{center}
		\label{tab:true_det}
	}
\end{table}

\newcommand{\fst}[1]{\footnotesize\textcolor{teal}{\bf #1}}
\newcommand{\snd}[1]{\footnotesize\textcolor{blue}{\bf #1}}
\newcommand{\fsta}[1]{\footnotesize\textcolor{teal}{#1}}
\newcommand{\snda}[1]{\footnotesize\textcolor{blue}{#1}}

\begin{table}[t]
{
	\caption{Performance comparison between TextBoxes++ (TB++), DAL, Oriented RCNN (OCNN) and FPSSD-RBox (FPRB) for the Visual Inspection task and different values of the confidence threshold $\tau_c$. (TP + FN = nGT [Table~\ref{tab:true_det}], \fst{green} and \snd{blue} resp. denote \fst{best} and \snd{2nd best} for every confidence threshold value.)}
	\footnotesize
	\begin{center}
	\begin{tabular}{lcccccccccc} 
		\toprule
		Method   & $\tau_c$ & TP     & FP           & FN           & R            & P            & F$_1$        &  AP          & ARIOU      \\
		\midrule                                                               
		\multirow{4}{1cm}{TB++}
		         & 0.5    & 592      & 74           & 130          & 0.8199       & \snd{0.8889} & \snd{0.8530} & 0.7954       & 0.4474     \\
		         & 0.6    & 549      & 154          & 173          & \snd{0.7604} & 0.7809       & 0.7705       & 0.6853       & 0.4712     \\
		         & 0.7    & 460      & 368          & 262          & 0.6371       & 0.5556       & 0.5935       & 0.4851       & 0.4615     \\
		         & 0.8    & 396      & 527          & 326          & 0.5485       & 0.4290       & 0.4815       & 0.3875       & 0.3612     \\
		\midrule                                                               
		\multirow{4}{1cm}{DAL}
		         & 0.5    & 592      & 155          & 130          & 0.8199       & 0.7925      & 0.8060        & 0.7731       & 0.6774           \\
		         & 0.6    & 526      & 221          & 196          & 0.7285       & 0.7041      & 0.7161        & 0.6943       & \fst{0.6816}     \\
		         & 0.7    & 372      & 375          & 350          & 0.5152       & 0.4980      & 0.5065        & 0.4485       & \fst{0.6586}     \\
	       	     & 0.8    & 141      & 606          & 581          & 0.1953       & 0.1888      & 0.1920        & 0.1475       & 0.5671           \\
		\midrule                                                               
		\multirow{4}{1cm}{OCNN}
		         & 0.5    & 604      & 94           & 118          & \snd{0.8366} & 0.8653       & 0.8507       & \snd{0.8254} & \fst{0.6931}     \\
		         & 0.6    & 542      & 106          & 180          & 0.7507       & \snd{0.8364} & \snd{0.7912} & \snd{0.7353} & \snd{0.6243}     \\
		         & 0.7    & 486      & 182          & 236          & \snd{0.6731} & \snd{0.7275} & \snd{0.6993} & \snd{0.6651} & 0.5130           \\
		         & 0.8    & 453      & 246          & 269          & \snd{0.6274} & \snd{0.6481} & \snd{0.6376} & \snd{0.5875} & 0.4143           \\
		\midrule                                                               
		\multirow{4}{1cm}{FPRB}
		         & 0.5    & 722      & 62          & 118           & \fst{1.0000} & \fst{0.9209} & \fst{0.9588} & \fst{0.8412} & \snd{0.5861}     \\
		         & 0.6    & 712      & 89          & 180           & \fst{0.9861} & \fst{0.8889} & \fst{0.9350} & \fst{0.8686} & 0.5677           \\
		         & 0.7    & 657      & 0           & 236           & \fst{0.9100} & \fst{1.0000} & \fst{0.9529} & \fst{0.9091} & \snd{0.5932}     \\
		         & 0.8    & 622      & 147         & 269           & \fst{0.8615} & \fst{0.8088} & \fst{0.8343} & \fst{0.7486} & \fst{0.5737}     \\
		\bottomrule
	\end{tabular}
	\end{center}
	\label{tab:comp_insp}
}
\end{table}

\begin{table}
	{
		\caption{Performance comparison between TextBoxes++ (TB++), DAL, Oriented RCNN (OCNN) and FPSSD-RBox (FPRB) for the Quality Control task and a confidence threshold $\tau_c = 0.5$  (TP + FN = nGT [Table~\ref{tab:true_det}], \fst{green} and \snd{blue} resp. denote \fst{best} and \snd{2nd best}.)}
		\footnotesize
		\begin{center}
			\begin{tabular}{lccccccccc} 
				\toprule
				Method & Class & TP & FP & FN & R & P & F$_1$ & AP & ARIOU    \\
				\midrule                                                               
				\multirow{6}{1cm}{TB++}
				& LAB & 367 & 50 & 23 & 0.9410 & 0.8801 & 0.9095 & 0.8641 & 0.4584 \\
				& BLU & 517 & 10 & 81 & 0.8645 & 0.9810 & 0.9191 & 0.8155 & 0.3259 \\
				& PIN & 228 & 42 & 39 & 0.8539 & 0.8444 & 0.8492 & 0.8351 & 0.2483 \\
				& BLA & 480 & 11 & 61 & 0.8872 & 0.9776 & 0.9302 & 0.7835 & 0.2852 \\
				& SEA & 140 & 31 & 13 & 0.9150 & 0.8187 & 0.8642 & 0.8105 & 0.5071 \\
				& IFI & 189 & 20 & 0 & 1.0000 & 0.9043 & 0.9497 & 0.9053 & 0.5527 \\
				\cmidrule{2-10}
				& AVG &  &  &  & 0.9103 & \snd{0.9010} & 0.9056 & 0.8217 & 0.3963 \\
				\midrule                                                               
				\multirow{6}{1cm}{DAL}
				& LAB & 374 & 55 & 16 & 0.9590 & 0.8718 & 0.9133 & 0.8631 & 0.6978 \\
				& BLU & 504 & 94 & 94 & 0.8428 & 0.8428 & 0.8428 & 0.8406 & 0.5724 \\
				& PIN & 252 & 44 & 15 & 0.9438 & 0.8514 & 0.8952 & 0.8371 & 0.6650 \\
				& BLA & 481 & 46 & 60 & 0.8891 & 0.9127 & 0.9007 & 0.8648 & 0.6341 \\
				& SEA & 152 & 21 & 1 & 0.9935 & 0.8786 & 0.9325 & 0.8561 & 0.7621 \\
				& IFI & 188 & 7 & 1 & 0.9947 & 0.9641 & 0.9792 & 0.9465 & 0.6738 \\
				\cmidrule{2-10}
				& AVG &  &  &  & \fst{0.9371} & 0.8869 & \snd{0.9113} & \fst{0.8680} & \snd{0.6675} \\
				\midrule                                                               
				\multirow{6}{1cm}{OCNN}
				& LAB & 385 & 11 & 5 & 0.9872 & 0.9722 & 0.9796 & 0.9352 & 0.7341 \\
				& BLU & 445 & 116 & 153 & 0.7441 & 0.7932 & 0.7679 & 0.7009 & 0.6324 \\
				& PIN & 255 & 68 & 12 & 0.9551 & 0.7895 & 0.8644 & 0.7646 & 0.6752 \\
				& BLA & 474 & 56 & 67 & 0.8762 & 0.8943 & 0.8852 & 0.8238 & 0.6691 \\
				& SEA & 150 & 15 & 3 & 0.9804 & 0.9091 & 0.9434 & 0.9064 & 0.7315 \\
				& IFI & 187 & 6 & 2 & 0.9894 & 0.9689 & 0.9791 & 0.9191 & 0.6642 \\
				\cmidrule{2-10}
				& AVG &  &  &  & 0.9221 & 0.8879 & 0.9046 & \snd{0.8417} & \fst{0.6844} \\
				\midrule                                                               
				\multirow{6}{1cm}{FPRB}
				& LAB & 374 & 59 & 16 & 0.9590 & 0.8637 & 0.9089 & 0.8471 & 0.7263 \\
				& BLU & 537 & 17 & 61 & 0.8980 & 0.9693 & 0.9323 & 0.7364 & 0.6374 \\
				& PIN & 222 & 37 & 45 & 0.8315 & 0.8571 & 0.8441 & 0.8184 & 0.5052 \\
				& BLA & 507 & 11 & 34 & 0.9372 & 0.9788 & 0.9575 & 0.8074 & 0.4371 \\
				& SEA & 148 & 27 & 5 & 0.9673 & 0.8457 & 0.9024 & 0.8174 & 0.6794 \\
				& IFI & 189 & 7 & 0 & 1.0000 & 0.9643 & 0.9818 & 0.9642 & 0.6636 \\
				\cmidrule{2-10}
				& AVG &  &  &  & \snd{0.9322} & \fst{0.9132} & \fst{0.9226} & 0.8318 & 0.6082 \\
				\bottomrule
			\end{tabular}
		\end{center}
		\label{tab:comp_qual_05}
	}
\end{table}

\begin{table}
	{
		\caption{Performance comparison between TextBoxes++ (TB++), DAL, Oriented RCNN (OCNN) and FPSSD-RBox (FPRB) for the Quality Control task and a confidence threshold $\tau_c = 0.6$  (TP + FN = nGT [Table~\ref{tab:true_det}], \fst{green} and \snd{blue} resp. denote \fst{best} and \snd{2nd best}.)}
		\footnotesize
		\begin{center}
			\begin{tabular}{lccccccccc} 
				\toprule
				Method   & Class & TP     & FP           & FN          & R       & P      & F$_1$ &   AP      & ARIOU    \\
				\midrule                                                               
				\multirow{6}{1cm}{TB++}
				& LAB & 366 & 38 & 24 & 0.9385 & 0.9059 & 0.9219 & 0.8904 & 0.4661 \\
				& BLU & 482 & 143 & 116 & 0.8060 & 0.7712 & 0.7882 & 0.7681 & 0.2963 \\
				& PIN & 220 & 30 & 47 & 0.8240 & 0.8800 & 0.8511 & 0.8005 & 0.2574 \\
				& BLA & 442 & 72 & 99 & 0.8170 & 0.8599 & 0.8379 & 0.8066 & 0.2730 \\
				& SEA & 132 & 21 & 21 & 0.8627 & 0.8627 & 0.8627 & 0.8570 & 0.5372 \\
				& IFI & 186 & 13 & 3 & 0.9841 & 0.9347 & 0.9588 & 0.9233 & 0.5517 \\
				\cmidrule{2-10}
				& AVG &  &  &  & 0.8721 & 0.8691 & 0.8706 & 0.8410 & 0.3970 \\
				\midrule                                                               
				\multirow{6}{1cm}{DAL}
				& LAB & 368 & 49 & 22 & 0.9436 & 0.8825 & 0.9120 & 0.8612 & 0.6854 \\
				& BLU & 459 & 12 & 139 & 0.7676 & 0.9745 & 0.8587 & 0.7514 & 0.6196 \\
				& PIN & 243 & 32 & 24 & 0.9101 & 0.8836 & 0.8967 & 0.8763 & 0.6448 \\
				& BLA & 446 & 81 & 95 & 0.8244 & 0.8463 & 0.8352 & 0.8064 & 0.6283 \\
				& SEA & 146 & 16 & 7 & 0.9542 & 0.9012 & 0.9270 & 0.8658 & 0.7495 \\
				& IFI & 187 & 3 & 2 & 0.9894 & 0.9842 & 0.9868 & 0.9785 & 0.6581 \\
				\cmidrule{2-10}
				& AVG &  &  &  & \snd{0.8982} & \fst{0.9121} & \fst{0.9051} & \fst{0.8566} & \fst{0.6643} \\
				\midrule                                                               
				\multirow{6}{1cm}{OCNN}
				& LAB & 375 & 10 & 15 & 0.9615 & 0.9740 & 0.9677 & 0.9067 & 0.7261 \\
				& BLU & 351 & 126 & 247 & 0.5870 & 0.7358 & 0.6530 & 0.5192 & 0.5802 \\
				& PIN & 236 & 65 & 31 & 0.8839 & 0.7841 & 0.8310 & 0.7738 & 0.6485 \\
				& BLA & 435 & 38 & 106 & 0.8041 & 0.9197 & 0.8580 & 0.7714 & 0.6311 \\
				& SEA & 144 & 13 & 9 & 0.9412 & 0.9172 & 0.9290 & 0.8962 & 0.6843 \\
				& IFI & 185 & 4 & 4 & 0.9788 & 0.9788 & 0.9788 & 0.9091 & 0.6385 \\
				\cmidrule{2-10}
				& AVG &  &  &  & 0.8594 & 0.8849 & 0.8720 & 0.7961 & \snd{0.6515} \\
				\midrule                                                               
				\multirow{6}{1cm}{FPRB}
				& LAB & 355 & 14 & 35 & 0.9103 & 0.9621 & 0.9354 & 0.8961 & 0.6941 \\
				& BLU & 526 & 148 & 72 & 0.8796 & 0.7804 & 0.8270 & 0.7763 & 0.6354 \\
				& PIN & 210 & 25 & 57 & 0.7865 & 0.8936 & 0.8367 & 0.7627 & 0.5385 \\
				& BLA & 500 & 91 & 41 & 0.9242 & 0.8460 & 0.8834 & 0.8361 & 0.4863 \\
				& SEA & 146 & 24 & 7 & 0.9542 & 0.8588 & 0.9040 & 0.8284 & 0.6837 \\
				& IFI & 189 & 2 & 0 & 1.0000 & 0.9895 & 0.9947 & 0.9878 & 0.6648 \\
				\cmidrule{2-10}
				& AVG &  &  &  & \fst{0.9091} & \snd{0.8884} & \snd{0.8987} & \snd{0.8479} & 0.6171 \\
				\bottomrule
			\end{tabular}
		\end{center}
		\label{tab:comp_qual_06}
	}
\end{table}

\begin{table}
	{
		\caption{Performance comparison between TextBoxes++ (TB++), DAL, Oriented RCNN (OCNN) and FPSSD-RBox (FPRB) for the Quality Control task and a confidence threshold $\tau_c = 0.7$  (TP + FN = nGT [Table~\ref{tab:true_det}], \fst{green} and \snd{blue} resp. denote \fst{best} and \snd{2nd best}.)}
		\footnotesize
		\begin{center}
			\begin{tabular}{lccccccccc} 
				\toprule
				Method   & Class & TP     & FP           & FN          & R       & P      & F$_1$ &   AP      & ARIOU    \\
				\midrule                                                               
				\multirow{6}{1cm}{TB++}
				& LAB & 352 & 45 & 38 & 0.9026 & 0.8866 & 0.8945 & 0.8515 & 0.4851 \\
				& BLU & 435 & 154 & 163 & 0.7274 & 0.7385 & 0.7329 & 0.7153 & 0.3197 \\
				& PIN & 216 & 31 & 51 & 0.8090 & 0.8745 & 0.8405 & 0.7938 & 0.2683 \\
				& BLA & 422 & 96 & 119 & 0.7800 & 0.8147 & 0.7970 & 0.7775 & 0.2957 \\
				& SEA & 127 & 27 & 26 & 0.8301 & 0.8247 & 0.8274 & 0.8167 & 0.5332 \\
				& IFI & 179 & 9 & 10 & 0.9471 & 0.9521 & 0.9496 & 0.9366 & 0.5712 \\
				\cmidrule{2-10}
				& AVG &  &  &  & 0.8327 & 0.8485 & 0.8405 & \snd{0.8152} & 0.4122 \\
				\midrule                                                               
				\multirow{6}{1cm}{DAL}
				& LAB & 349 & 60 & 41 & 0.8949 & 0.8533 & 0.8736 & 0.8336 & 0.5674 \\
				& BLU & 427 & 161 & 171 & 0.7140 & 0.7262 & 0.7201 & 0.7053 & 0.6354 \\
				& PIN & 226 & 47 & 41 & 0.8464 & 0.8278 & 0.8370 & 0.8050 & 0.5875 \\
				& BLA & 412 & 143 & 129 & 0.7616 & 0.7423 & 0.7518 & 0.7341 & 0.6264 \\
				& SEA & 137 & 28 & 16 & 0.8954 & 0.8303 & 0.8616 & 0.8157 & 0.6875 \\
				& IFI & 186 & 16 & 3 & 0.9841 & 0.9208 & 0.9514 & 0.8963 & 0.6471 \\
				\cmidrule{2-10}
				& AVG &  &  &  & \snd{0.8494} & 0.8168 & 0.8328 & 0.7983 & 0.6252 \\
				\midrule                                                               
				\multirow{6}{1cm}{OCNN}
				& LAB & 368 & 10 & 22 & 0.9436 & 0.9735 & 0.9583 & 0.8873 & 0.6943 \\
				& BLU & 314 & 111 & 284 & 0.5251 & 0.7388 & 0.6139 & 0.4857 & 0.5943 \\
				& PIN & 227 & 52 & 40 & 0.8502 & 0.8136 & 0.8315 & 0.7810 & 0.6274 \\
				& BLA & 411 & 35 & 130 & 0.7597 & 0.9215 & 0.8328 & 0.7338 & 0.6531 \\
				& SEA & 140 & 11 & 13 & 0.9150 & 0.9272 & 0.9211 & 0.8804 & 0.6411 \\
				& IFI & 183 & 10 & 6 & 0.9683 & 0.9482 & 0.9581 & 0.8967 & 0.6474 \\
				\cmidrule{2-10}
				& AVG &  &  &  & 0.8270 & \snd{0.8871} & \snd{0.8560} & 0.7775 & \snd{0.6429} \\
				\midrule                                                               
				\multirow{6}{1cm}{FPRB}
				& LAB & 358 & 8 & 32 & 0.9179 & 0.9781 & 0.9471 & 0.9097 & 0.7102 \\
				& BLU & 430 & 59 & 168 & 0.7191 & 0.8793 & 0.7912 & 0.7055 & 0.6247 \\
				& PIN & 244 & 15 & 23 & 0.9139 & 0.9421 & 0.9278 & 0.9093 & 0.5604 \\
				& BLA & 445 & 18 & 96 & 0.8226 & 0.9611 & 0.8865 & 0.8206 & 0.4993 \\
				& SEA & 131 & 4 & 22 & 0.8562 & 0.9704 & 0.9097 & 0.8461 & 0.7123 \\
				& IFI & 189 & 0 & 0 & 1.0000 & 1.0000 & 1.0000 & 1.0000 & 0.7669 \\
				\cmidrule{2-10}
				& AVG &  &  &  & \fst{0.8716} & \fst{0.9552} & \fst{0.9115} & \fst{0.8652} & \fst{0.6456} \\
				\bottomrule
			\end{tabular}
		\end{center}
		\label{tab:comp_qual_07}
	}
\end{table}

\begin{table}
	{
		\caption{Performance comparison between TextBoxes++ (TB++), DAL, Oriented RCNN (OCNN) and FPSSD-RBox (FPRB) for the Quality Control task and a confidence threshold $\tau_c = 0.8$  (TP + FN = nGT [Table~\ref{tab:true_det}], \fst{green} and \snd{blue} resp. denote \fst{best} and \snd{2nd best}.)}
		\footnotesize
		\begin{center}
			\begin{tabular}{lccccccccc} 
				\toprule
				Method   & Class & TP     & FP           & FN          & R       & P      & F$_1$ &   AP      & ARIOU    \\
				\midrule                                                               
				\multirow{6}{1cm}{TB++}
				& LAB & 336 & 86 & 54 & 0.8615 & 0.7962 & 0.8276 & 0.7756 & 0.4374 \\
				& BLU & 428 & 18 & 170 & 0.7157 & 0.9596 & 0.8199 & 0.6851 & 0.3081 \\
				& PIN & 205 & 67 & 62 & 0.7678 & 0.7537 & 0.7607 & 0.7417 & 0.2371 \\
				& BLA & 377 & 14 & 164 & 0.6969 & 0.9642 & 0.8090 & 0.6861 & 0.2851 \\
				& SEA & 120 & 33 & 33 & 0.7843 & 0.7843 & 0.7843 & 0.7811 & 0.5449 \\
				& IFI & 177 & 11 & 12 & 0.9365 & 0.9415 & 0.9390 & 0.9264 & 0.5722 \\
				\cmidrule{2-10}
				& AVG &  &  &  & \fst{0.7938} & \snd{0.8666} & \fst{0.8286} & \fst{0.7660} & 0.3975 \\
				\midrule                                                               
				\multirow{6}{1cm}{DAL}
				& LAB & 323 & 59 & 67 & 0.8282 & 0.8455 & 0.8368 & 0.8164 & 0.5554 \\
				& BLU & 401 & 18 & 197 & 0.6706 & 0.9570 & 0.7886 & 0.6677 & 0.5996 \\
				& PIN & 210 & 53 & 57 & 0.7865 & 0.7985 & 0.7925 & 0.7753 & 0.6052 \\
				& BLA & 355 & 17 & 186 & 0.6562 & 0.9543 & 0.7777 & 0.6471 & 0.5963 \\
				& SEA & 121 & 36 & 32 & 0.7908 & 0.7707 & 0.7806 & 0.7518 & 0.6861 \\
				& IFI & 175 & 15 & 14 & 0.9259 & 0.9211 & 0.9235 & 0.9054 & 0.6474 \\
				\cmidrule{2-10}
				& AVG &  &  &  & 0.7764 & \fst{0.8745} & 0.8225 & \snd{0.7606} & \fst{0.6150} \\
				\midrule                                                               
				\multirow{6}{1cm}{OCNN}
				& LAB & 322 & 16 & 68 & 0.8256 & 0.9527 & 0.8846 & 0.8124 & 0.6153 \\
				& BLU & 247 & 164 & 351 & 0.4130 & 0.6010 & 0.4896 & 0.4042 & 0.5572 \\
				& PIN & 207 & 47 & 60 & 0.7753 & 0.8150 & 0.7946 & 0.7631 & 0.5489 \\
				& BLA & 354 & 47 & 187 & 0.6543 & 0.8828 & 0.7516 & 0.6258 & 0.5722 \\
				& SEA & 116 & 17 & 37 & 0.7582 & 0.8722 & 0.8112 & 0.7368 & 0.6149 \\
				& IFI & 178 & 13 & 11 & 0.9418 & 0.9319 & 0.9368 & 0.8816 & 0.5953 \\
				\cmidrule{2-10}
				& AVG &  &  &  & 0.7280 & 0.8426 & 0.7811 & 0.7040 & 0.5840 \\
				\midrule                                                               
				\multirow{6}{1cm}{FPRB}
				& LAB & 330 & 67 & 60 & 0.8462 & 0.8312 & 0.8386 & 0.8274 & 0.6671 \\
				& BLU & 430 & 18 & 168 & 0.7191 & 0.9598 & 0.8222 & 0.6752 & 0.6081 \\
				& PIN & 207 & 58 & 60 & 0.7753 & 0.7811 & 0.7782 & 0.7682 & 0.5465 \\
				& BLA & 369 & 20 & 172 & 0.6821 & 0.9486 & 0.7935 & 0.6418 & 0.4762 \\
				& SEA & 119 & 32 & 34 & 0.7778 & 0.7881 & 0.7829 & 0.7692 & 0.6853 \\
				& IFI & 170 & 22 & 19 & 0.8995 & 0.8854 & 0.8924 & 0.8741 & 0.6515 \\
				\cmidrule{2-10}
				& AVG &  &  &  & \snd{0.7833} & 0.8657 & \snd{0.8226} & 0.7593 & \snd{0.6058} \\
				\bottomrule
			\end{tabular}
		\end{center}
		\label{tab:comp_qual_08}
	}
\end{table}

\begin{table}
{
	\caption{Best performance achieved regarding the detection of the different classes involved in the Visual Inspection (VI) and Quality Control (QC) tasks: every cell indicates the \fsta{best-} and \snda{second-best} performing algorithms, the metric values and the corresponding confidence thresholds $\tau_c$.}
	\footnotesize
	\begin{center}
	\begin{tabular}{@{\hspace{0mm}}c@{\hspace{2mm}}c@{\hspace{2mm}}c@{\hspace{2mm}}c@{\hspace{2mm}}c@{\hspace{2mm}}c@{\hspace{2mm}}c@{\hspace{0mm}}} 
	\toprule
	Task   & Class & R                 & P                 & F$_1$             &   AP              & ARIOU        \\
	\midrule
	\bf VI     
	       & COR   & \fsta{FPRB} & \fsta{FPRB} & \fsta{FPRB} & \fsta{FPRB} & \fsta{OCNN}  \\
	       &       & 1.0000 (0.5)      & 1.0000 (0.7)      & 0.9588 (0.5)      & 0.9091 (0.7)      & 0.6931 (0.5) \\
	       &       & \snda{FPRB} & \snda{FPRB} & \snda{FPRB} & \snda{FPRB} & \snda{DAL}   \\
	       &       & 0.9861 (0.6)      & 0.9209 (0.5)      & 0.9529 (0.7)      & 0.8686 (0.6)      & 0.6816 (0.6) \\
	\midrule
	\bf QC     
	       & LAB   & \fsta{OCNN}       & \fsta{FPRB} & \fsta{OCNN}       & \fsta{OCNN}       & \fsta{OCNN}  \\
	       &       & 0.9872 (0.5)      & 0.9781 (0.7)      & 0.9796 (0.5)      & 0.9352 (0.5)      & 0.7341 (0.5) \\
	       &       & \snda{OCNN}       & \snda{OCNN}       & \snda{OCNN}       & \snda{FPRB} & \snda{FPRB} \\
	       &       & 0.9615 (0.6)      & 0.9740 (0.6)      & 0.9677 (0.7)      & 0.9097 (0.7)      & 0.7263 (0.5) \\
	       \cmidrule{2-7}
	       & BLU   & \fsta{FPRB} & \fsta{TB++}       & \fsta{FPRB} & \fsta{DAL}        & \fsta{FPRB} \\
	       &       & 0.8980 (0.5)      & 0.9810 (0.5)      & 0.9323 (0.5)      & 0.8406 (0.5)      & 0.6374 (0.5) \\
	       &       & \snda{FPRB} & \snda{DAL}        & \snda{TB++}       & \snda{TB++}       & \snda{FPRB} \\
	       &       & 0.8796 (0.6)      & 0.9745 (0.6)      & 0.9191 (0.5)      & 0.8155 (0.5)      & 0.6354 (0.6) \\
	       \cmidrule{2-7}
	       & PIN   & \fsta{OCNN}       & \fsta{FPRB} & \fsta{FPRB} & \fsta{FPRB} & \fsta{OCNN}  \\
	       &       & 0.9551 (0.5)      & 0.9421 (0.7)      & 0.9278 (0.7)      & 0.9093 (0.7)      & 0.6752 (0.5) \\
	       &       & \snda{DAL}        & \snda{FPRB} & \snda{DAL}        & \snda{DAL}        & \snda{DAL}   \\
	       &       & 0.9438 (0.5)      & 0.8936 (0.6)      & 0.8967 (0.6)      & 0.8763 (0.6)      & 0.6650 (0.5) \\
	       \cmidrule{2-7}
	       & BLA   & \fsta{FPRB} & \fsta{FPRB} & \fsta{FPRB} & \fsta{DAL}        & \fsta{OCNN}  \\
	       &       & 0.9372 (0.5)      & 0.9788 (0.5)      & 0.9575 (0.5)      & 0.8648 (0.5)      & 0.6691 (0.5) \\
	       &       & \snda{FPRB} & \snda{TB++}       & \snda{TB++}       & \snda{FPRB} & \snda{OCNN}  \\
	       &       & 0.9242 (0.6)      & 0.9776 (0.5)      & 0.9302 (0.5)      & 0.8361 (0.6)      & 0.6531 (0.7) \\
	       \cmidrule{2-7}
	       & SEA   & \fsta{DAL}        & \fsta{FPRB} & \fsta{OCNN}       & \fsta{OCNN}       & \fsta{DAL}   \\
	       &       & 0.9935 (0.5)      & 0.9704 (0.7)      & 0.9434 (0.5)      & 0.9064 (0.5)      & 0.7621 (0.5) \\
	       &       & \snda{FPRB} & \snda{OCNN}       & \snda{DAL}        & \snda{OCNN}       & \snda{DAL}   \\
	       &       & 0.9804 (0.5)      & 0.9272 (0.7)      & 0.9325 (0.5)      & 0.8962 (0.6)      & 0.7495 (0.6) \\
	       \cmidrule{2-7}
	       & IFI   & \fsta{FPRB} & \fsta{FPRB} & \fsta{FPRB} & \fsta{FPRB} & \fsta{FPRB} \\
	       &       & 1.0000 (0.5-0.7)  & 1.0000 (0.7)      & 1.0000 (0.7)      & 1.0000 (0.7)      & 0.7669 (0.7) \\
	       &       & \fsta{TB++}       & \snda{FPRB} & \snda{FPRB} & \snda{FPRB} & \snda{DAL}   \\
	       &       & 1.0000 (0.5)      & 0.9895 (0.6)      & 0.9947 (0.6)      & 0.9878 (0.6)      & 0.6738 (0.5) \\
	       \cmidrule{2-7}
		   & AVG   & \fsta{DAL}        & \fsta{FPRB}       & \fsta{FPRB}       & \fsta{DAL}        & \fsta{OCNN}  \\
		   &       & 0.9371 (0.5)      & 0.9552 (0.7)      & 0.9226 (0.5)      & 0.8680 (0.7)      & 0.6844 (0.5) \\
		   &       & \snda{FPRB}       & \snda{FPRB}       & \snda{FPRB}       & \snda{FPRB} & \snda{DAL}   \\
		   &       & 0.9322 (0.5)      & 0.9132 (0.5)      & 0.9115 (0.7)      & 0.8652 (0.7)      & 0.6675 (0.5) \\
	\bottomrule
	\end{tabular}
	\end{center}
	\label{tab:best1}	
}
\end{table}

\begin{table}
{
	\caption{Number of times TextBoxes++ (TB++), DAL, Oriented RCNN (OCNN) and FPSSD-RBox (FPRB) scored best or second-best [respectively, rows labeled as \textit{1st} and rows labeled as \textit{2nd}; the sum of both is reported in the upper row]. (ARI stands for ARIOU; highest values are highlighted in bold face.)}
	\scriptsize
	\begin{center}
		\begin{tabular}{l@{\hspace{0mm}}p{1mm}  
				c@{\hspace{2mm}}c@{\hspace{2mm}}c@{\hspace{2mm}}c@{\hspace{2mm}}c@{\hspace{2mm}} l@{\hspace{1mm}}p{1mm}  c@{\hspace{2mm}}c@{\hspace{2mm}}c@{\hspace{2mm}}c@{\hspace{2mm}}c@{\hspace{2mm}} l@{\hspace{1mm}}p{1mm}  c@{\hspace{2mm}}c@{\hspace{2mm}}c@{\hspace{2mm}}c@{\hspace{2mm}}c@{\hspace{2mm}} l}
			\toprule
			     && \multicolumn{13}{l}{Quality Control} & & \multicolumn{6}{l}{} \\
			\cmidrule{3-15} 
			     && \multicolumn{6}{l}{Each class independently} &
			      & \multicolumn{6}{l}{Average} &
			      & \multicolumn{6}{l}{Visual Inspection} \\
			\cmidrule{3-8} \cmidrule{10-15} \cmidrule{17-22}
			     && R & P & F$_1$ &  AP & ARI & & & R & P & F$_1$ &  AP & ARI & & & R & P & F$_1$ &  AP & ARI &  \\
			\cmidrule{1-1} \cmidrule{3-8} \cmidrule{10-15} \cmidrule{17-22}
		\bf	TB++ &&  1 & 2 & 2 & 1 & 0 & =\,6  &&  0 & 0 & 0 & 0 & 0 & =\,0  &&  0 & 0 & 0 & 0 & 0 & =\,0 \\
		\ \	1st  &&  1 & 1 & 1 & 0 & 0 & =\,3  &&  0 & 0 & 0 & 0 & 0 & =\,0  &&  0 & 0 & 0 & 0 & 0 & =\,0 \\
		\ \	2nd  &&  0 & 1 & 1 & 1 & 0 & =\,3  &&  0 & 0 & 0 & 0 & 0 & =\,0  &&  0 & 0 & 0 & 0 & 0 & =\,0 \\ 
			\cmidrule{1-1} \cmidrule{3-8} \cmidrule{10-15} \cmidrule{17-22}
		\bf	DAL  &&  2 & 1 & 2 & 3 & 4 & =\,12 &&  1 & 0 & 1 & 1 & 1 & =\,4  &&  0 & 0 & 0 & 0 & 1 & =\,1 \\
		\ \	1st  &&  1 & 1 & 0 & 2 & 1 & =\,5  &&  1 & 0 & 0 & 1 & 0 & =\,\bf 2  &&  0 & 0 & 0 & 0 & 0 & =\,0 \\
		\ \	2nd  &&  1 & 0 & 2 & 1 & 3 & =\,7  &&  0 & 0 & 1 & 0 & 1 & =\,2  &&  0 & 0 & 0 & 0 & 1 & =\,1 \\ 
			\cmidrule{1-1} \cmidrule{3-8} \cmidrule{10-15} \cmidrule{17-22}
		\bf	OCNN &&  4 & 2 & 3 & 3 & 4 & =\,16 &&  0 & 0 & 0 & 0 & 1 & =\,1  &&  0 & 0 & 0 & 0 & 1 & =\,1 \\
		\ \	1st  &&  2 & 0 & 2 & 2 & 3 & =\,9  &&  0 & 0 & 0 & 0 & 1 & =\,1  &&  0 & 0 & 0 & 0 & 1 & =\,1 \\
		\ \	2nd  &&  2 & 2 & 1 & 1 & 1 & =\,7  &&  0 & 0 & 0 & 0 & 0 & =\,0  &&  0 & 0 & 0 & 0 & 0 & =\,0 \\ 
			\cmidrule{1-1} \cmidrule{3-8} \cmidrule{10-15} \cmidrule{17-22}
		\bf	FPRB &&  7 & 7 & 5 & 5 & 4 & =\,\bf 28 &&  1 & 2 & 1 & 1 & 0 & =\,\bf 5  &&  2 & 2 & 2 & 2 & 0 & =\,\bf 8 \\
		\ \	1st  &&  5 & 5 & 4 & 2 & 2 & =\,\bf 18 &&  0 & 1 & 1 & 0 & 0 & =\,\bf 2  &&  1 & 1 & 1 & 1 & 0 & =\,\bf 4 \\
		\ \	2nd  &&  2 & 2 & 1 & 3 & 2 & =\,\bf 10 &&  1 & 1 & 0 & 1 & 0 & =\,\bf 3  &&  1 & 1 & 1 & 1 & 0 & =\,\bf 4 \\ 
			\bottomrule
		\end{tabular}
	\end{center}
	\label{tab:best2}	
}
\end{table}

\begin{table}
{
\caption{Average inference times per image (in milliseconds) and FPS for TextBoxes++ (TB++), DAL, Oriented RCNN (OCNN) and FPSSD-RBox (FPRB) for the visual inspection and quality control tasks.}
\footnotesize
\begin{center}
\begin{tabular}{lcccc}
	\toprule
	           & TB++ & DAL & OCNN & FPRB \\ 
	\midrule
	time/image (fps) & 24ms (41.67) & 33ms (30.30) & 66ms (15.15) & 18ms (55.56) \\
	\bottomrule
\end{tabular}
\end{center}
\label{tab:times}	
}
\end{table}

\begin{figure*}
	\begin{center}
    \begin{tabular}{@{\hspace{0mm}}c@{\hspace{1mm}}c@{\hspace{1mm}}c@{\hspace{1mm}}c@{\hspace{1mm}}c@{\hspace{0mm}}}
        \footnotesize (a) FPSSD-RBox &
        \footnotesize (b) TextBoxes++ &
        \footnotesize (c) DAL & 
		\footnotesize (d) Oriented RCNN
        \\
        \includegraphics[width=0.24\columnwidth,height=0.24\columnwidth]{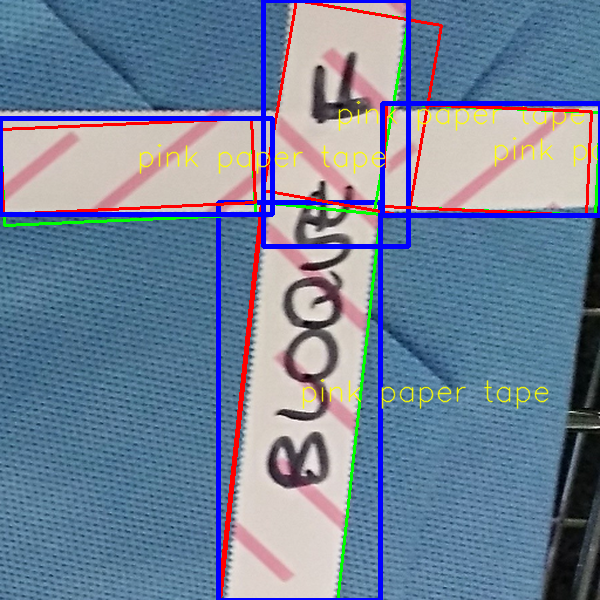}
        &
        \includegraphics[width=0.24\columnwidth,height=0.24\columnwidth]{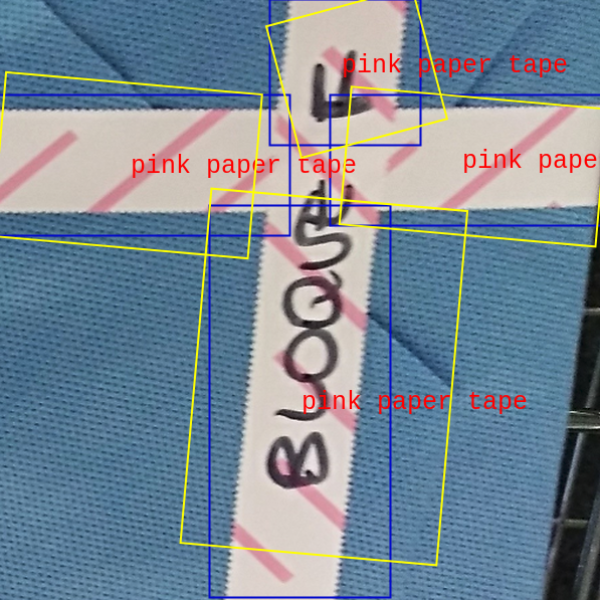}
        &
		\includegraphics[width=0.24\columnwidth,height=0.24\columnwidth]{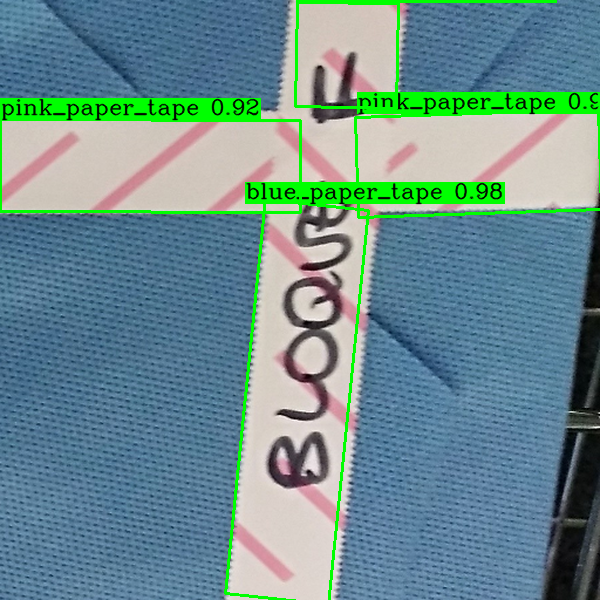}
        &
		\includegraphics[width=0.24\columnwidth,height=0.24\columnwidth]{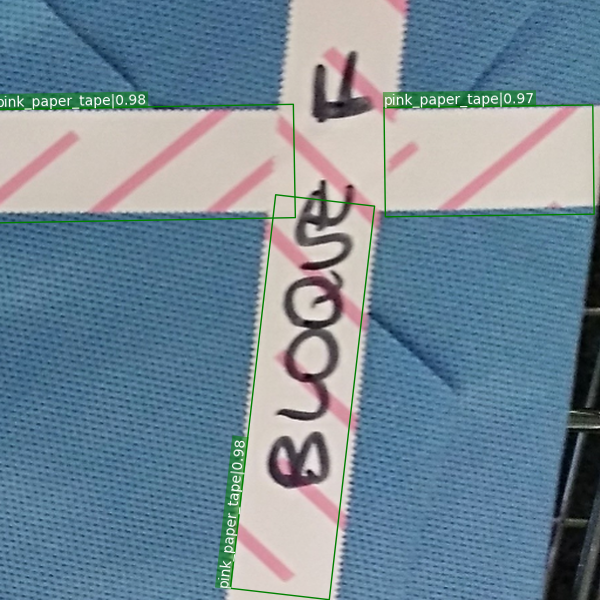}
        \\
        \includegraphics[width=0.24\columnwidth,height=0.24\columnwidth]{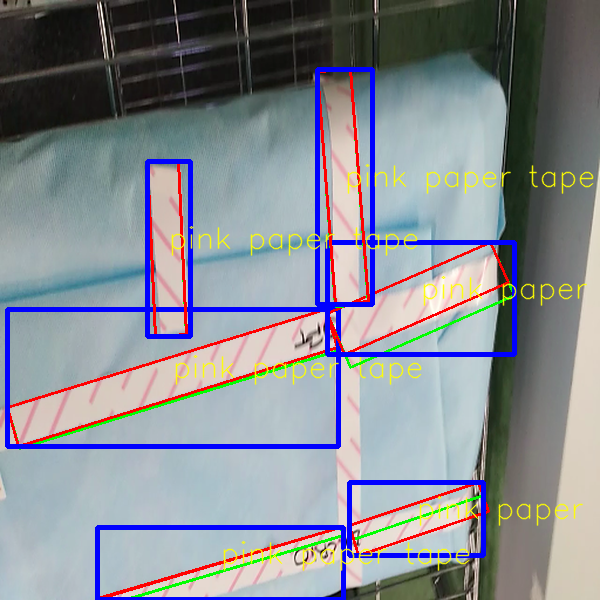}
        &
        \includegraphics[width=0.24\columnwidth,height=0.24\columnwidth]{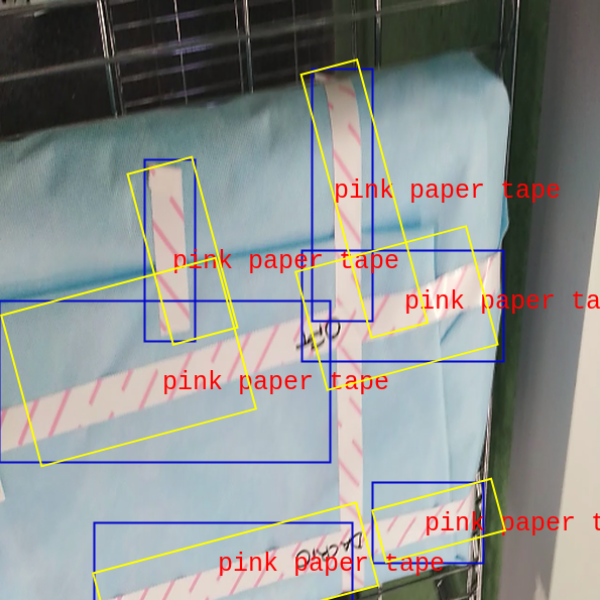}
        &
		\includegraphics[width=0.24\columnwidth,height=0.24\columnwidth]{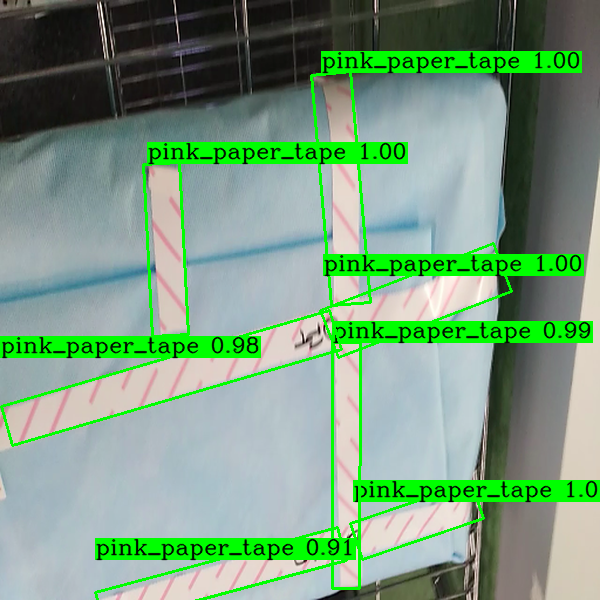}
        &
		\includegraphics[width=0.24\columnwidth,height=0.24\columnwidth]{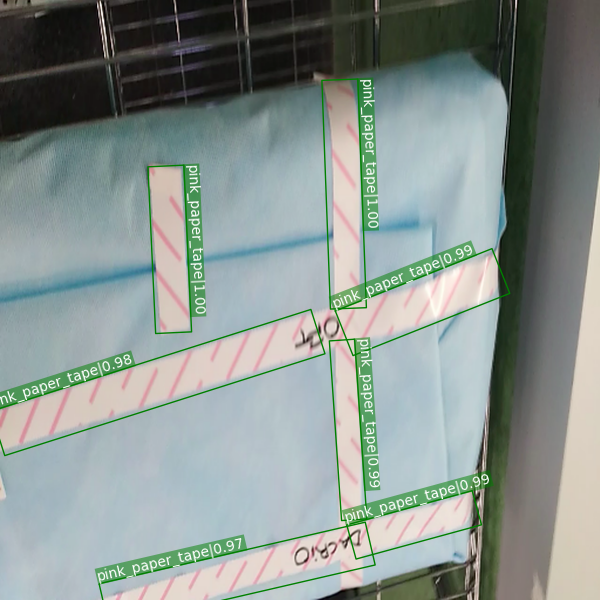}
        \\
        \includegraphics[width=0.24\columnwidth,height=0.24\columnwidth]{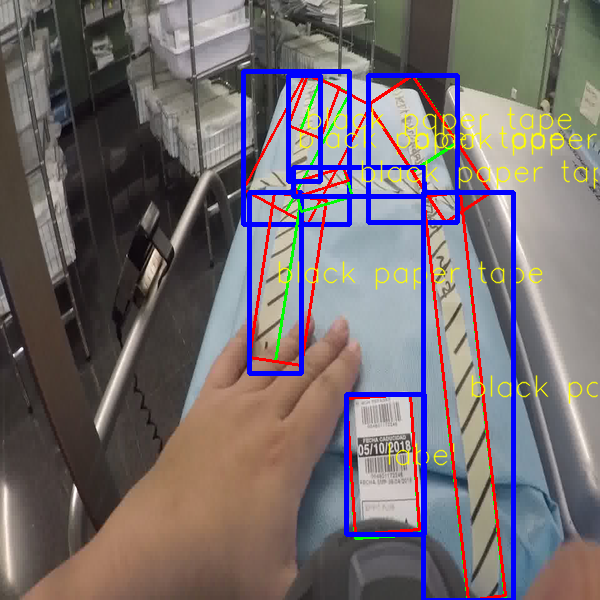}
        &
        \includegraphics[width=0.24\columnwidth,height=0.24\columnwidth]{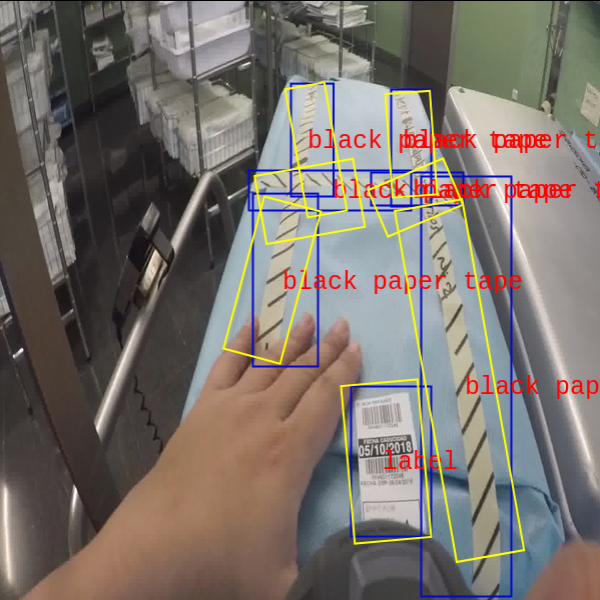}
        &
		\includegraphics[width=0.24\columnwidth,height=0.24\columnwidth]{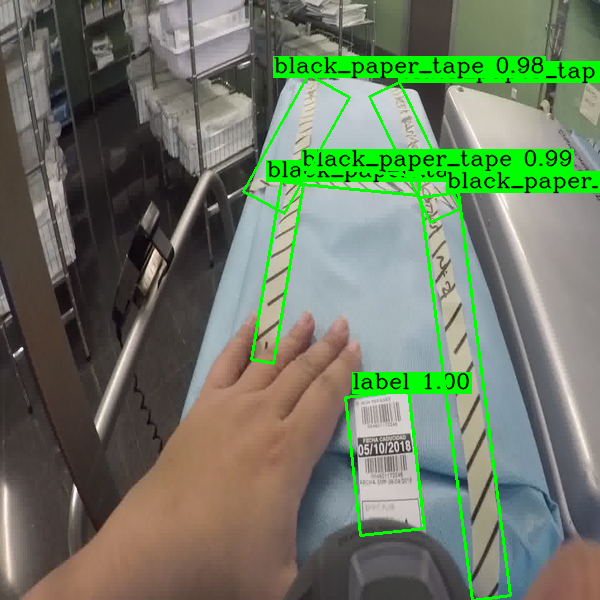}
        &
		\includegraphics[width=0.24\columnwidth,height=0.24\columnwidth]{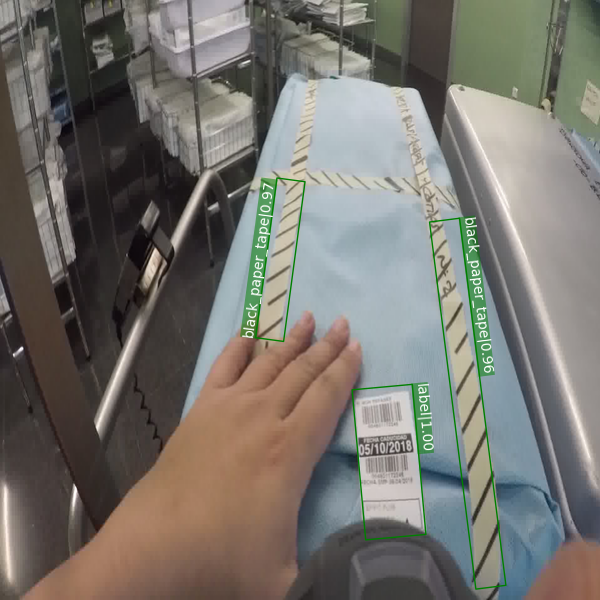}
        \\
        \includegraphics[width=0.24\columnwidth,height=0.24\columnwidth]{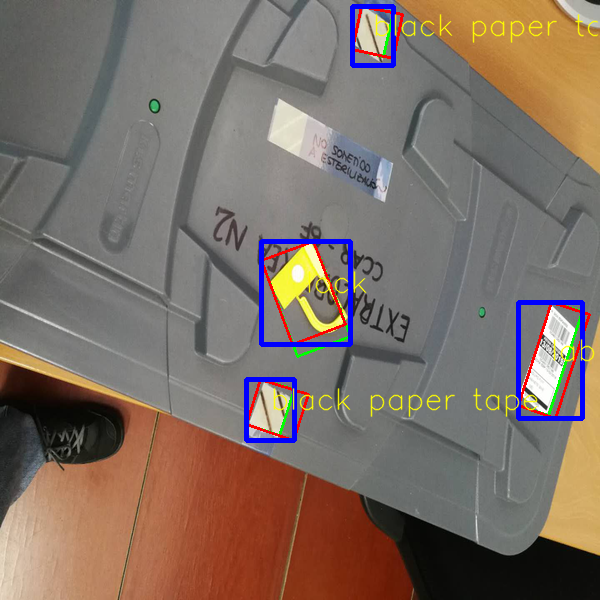}
        &
        \includegraphics[width=0.24\columnwidth,height=0.24\columnwidth]{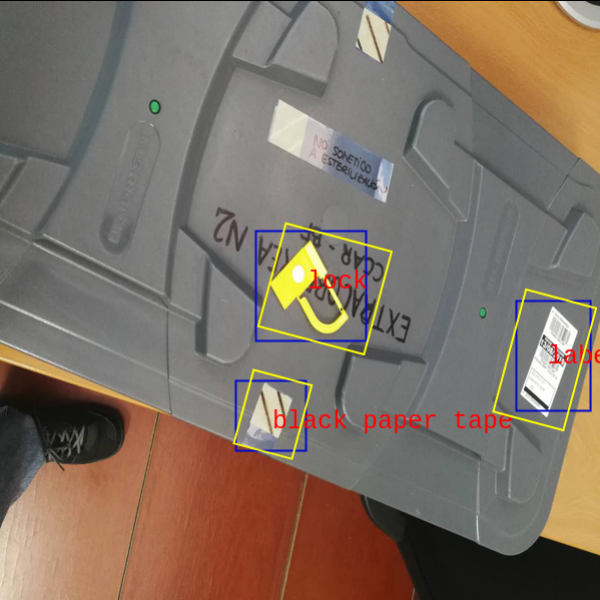}
        &
		\includegraphics[width=0.24\columnwidth,height=0.24\columnwidth]{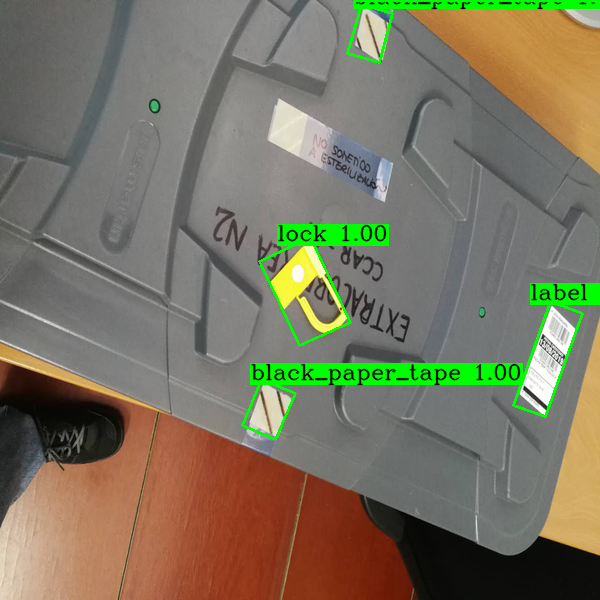}
        &
		\includegraphics[width=0.24\columnwidth,height=0.24\columnwidth]{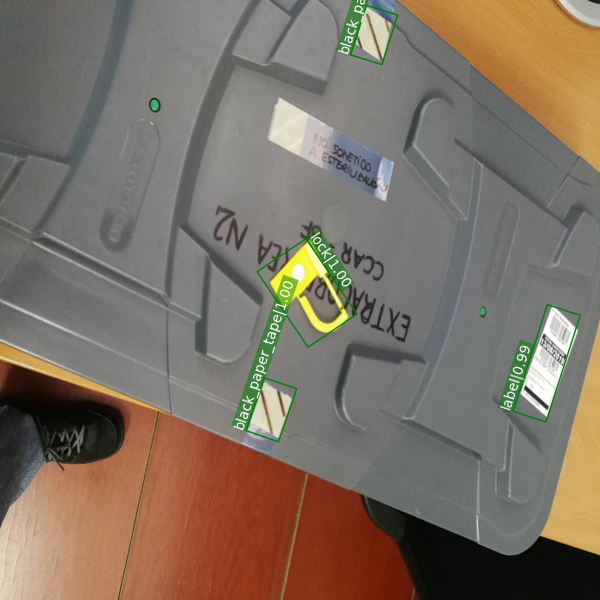}
    \end{tabular}
	\end{center}
    \vspace{-2mm}
    \caption{Examples of oriented detections for FPSSD-RBox, TextBoxes++, DAL and Oriented RCNN for the Quality Control task.}
    \label{fig:rbox_results1}
\end{figure*}
\begin{figure*}
    \begin{center}
    \begin{tabular}{@{\hspace{0mm}}c@{\hspace{1mm}}c@{\hspace{1mm}}c@{\hspace{1mm}}c@{\hspace{1mm}}c@{\hspace{0mm}}}
        \footnotesize (a) FPSSD-RBox &
        \footnotesize (b) TextBoxes++ &
        \footnotesize (c) DAL & 
        \footnotesize (d) Oriented RCNN
        \\
        \includegraphics[width=0.24\columnwidth,height=0.24\columnwidth]{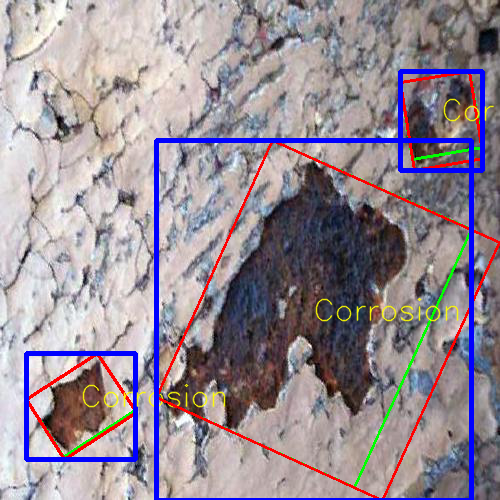}
        &
        \includegraphics[width=0.24\columnwidth,height=0.24\columnwidth]{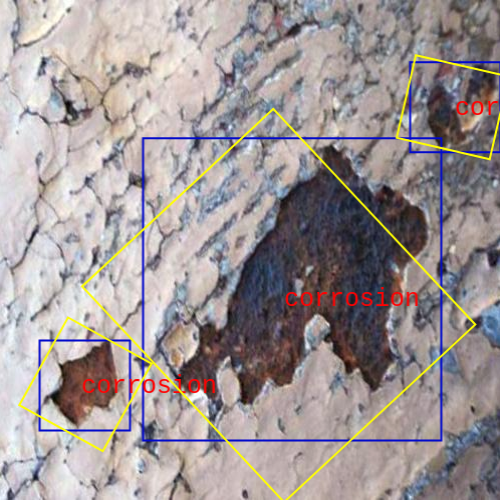}
        &
        \includegraphics[width=0.24\columnwidth,height=0.24\columnwidth]{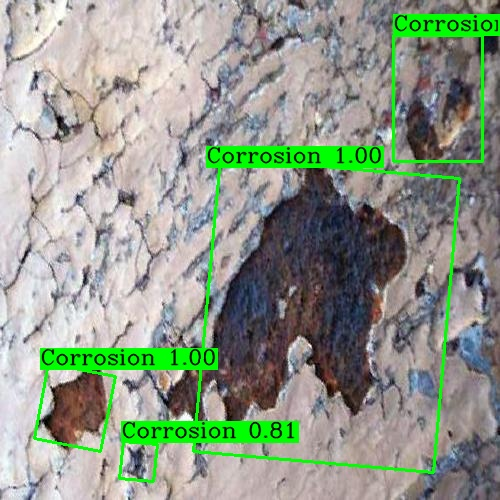}
        &
		\includegraphics[width=0.24\columnwidth,height=0.24\columnwidth]{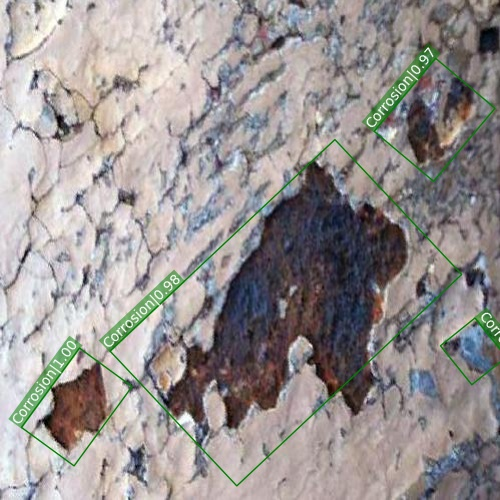}
        \\
        \includegraphics[width=0.24\columnwidth,height=0.24\columnwidth]{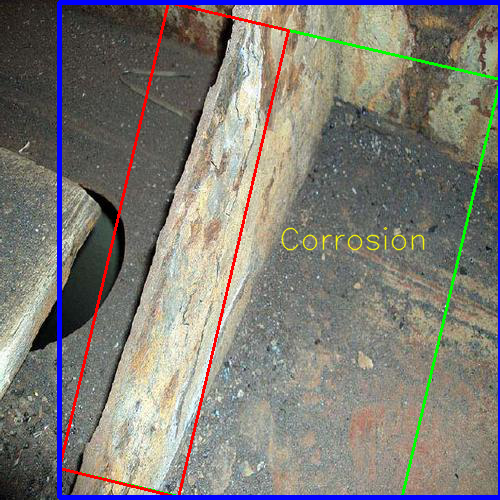}
        &
        \includegraphics[width=0.24\columnwidth,height=0.24\columnwidth]{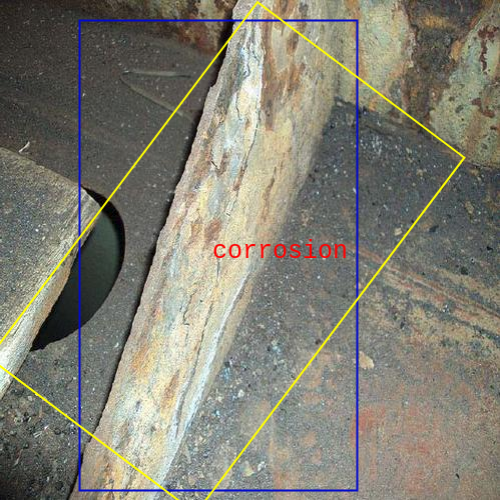}
        &
		\includegraphics[width=0.24\columnwidth,height=0.24\columnwidth]{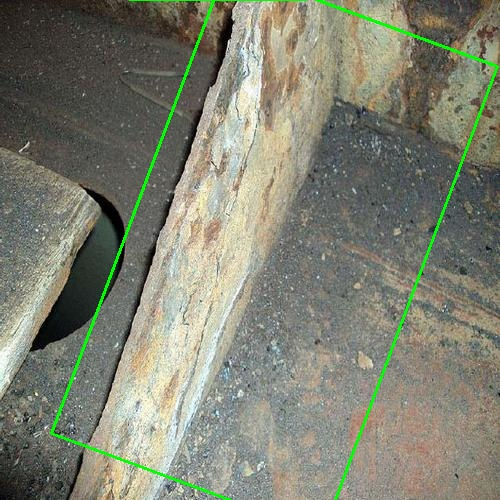}
        &
		\includegraphics[width=0.24\columnwidth,height=0.24\columnwidth]{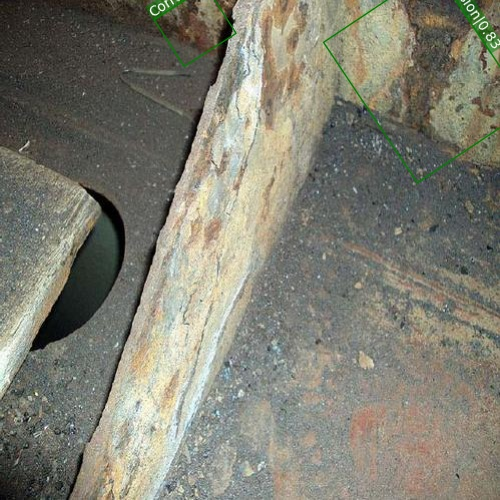}
        \\
        \includegraphics[width=0.24\columnwidth,height=0.24\columnwidth]{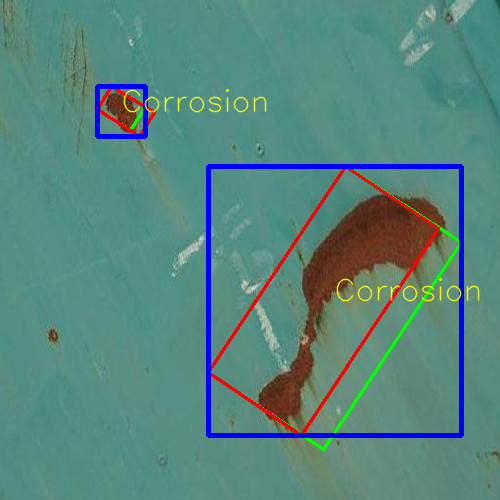}
        &
        \includegraphics[width=0.24\columnwidth,height=0.24\columnwidth]{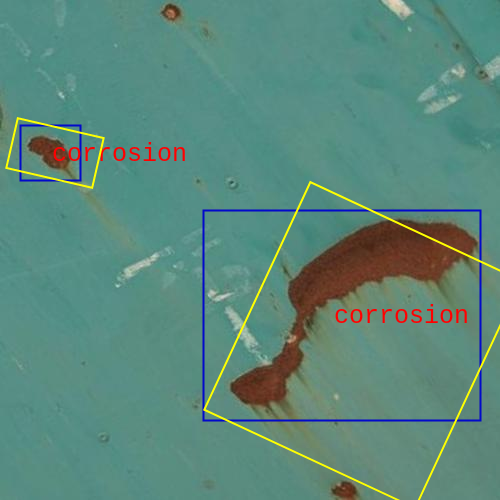}
        &
		\includegraphics[width=0.24\columnwidth,height=0.24\columnwidth]{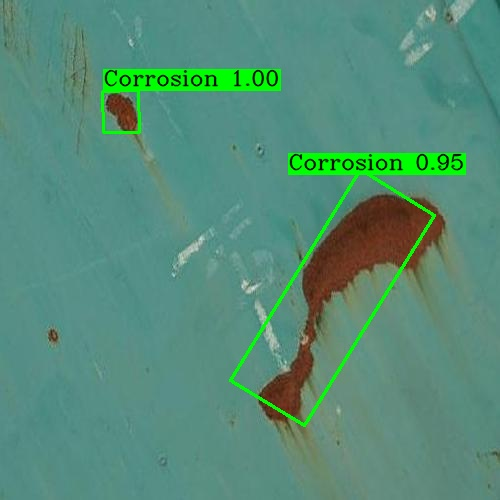}
        &
		\includegraphics[width=0.24\columnwidth,height=0.24\columnwidth]{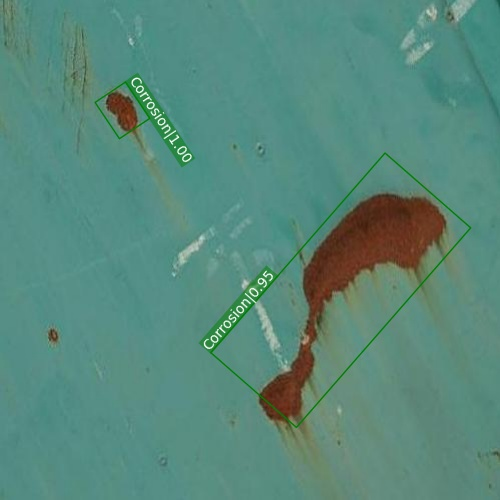}
        \\
        \includegraphics[width=0.24\columnwidth,height=0.24\columnwidth]{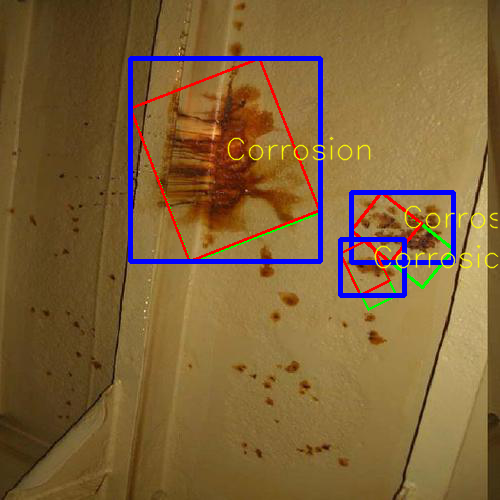}
        &
        \includegraphics[width=0.24\columnwidth,height=0.24\columnwidth]{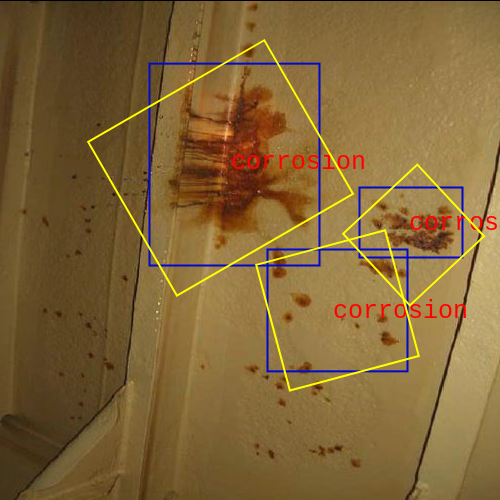}
        &
		\includegraphics[width=0.24\columnwidth,height=0.24\columnwidth]{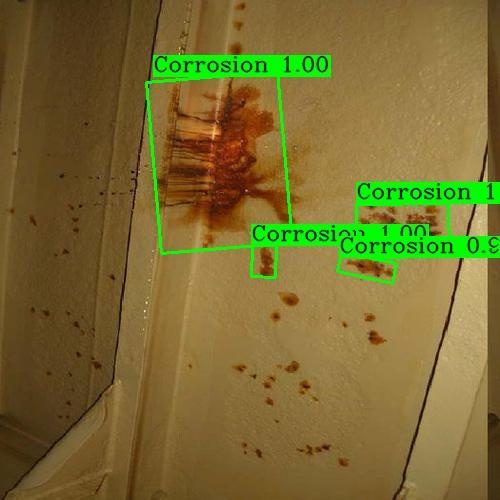}
        &
		\includegraphics[width=0.24\columnwidth,height=0.24\columnwidth]{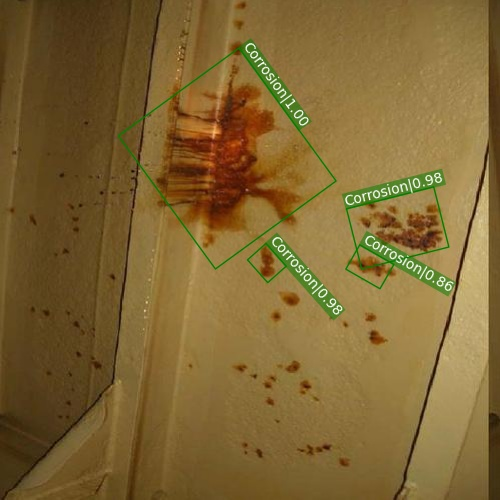}
    \end{tabular}
	\end{center}
    \vspace{-2mm}
    \caption{Examples of oriented detections for FPSSD-RBox, TextBoxes++, DAL and Oriented RCNN for the Visual Inspection task.}
    \label{fig:rbox_results2}
\end{figure*}

{To finish, Fig.~\ref{fig:rbox_results1} and~\ref{fig:rbox_results2} show superimposed detection results from FPSSD-RBox, TextBoxes++, DAL and Oriented RCNN for, respectively, the quality control task and the visual inspection task. These qualitative results show the same kind of performance that has been observed at the quantitative level: FPSSD-RBox gives rise to accurate predictions with a low number of FP/FN, similarly to DAL and Oriented RCNN, also providing reasonable predictions, though occasionally missing some detections (e.g. see Fig.~\ref{fig:rbox_results1}(d), first and third rows), while TextBoxes++ produces the least accurate predictions.}

\section{Conclusions}
\label{sc:conclusions}

A two-stage arbitrarily-oriented object detection method for regressing the parameters of oriented bounding boxes has been described, and assessed on two substantially different tasks. The first stage of our solution comprises a feature pyramid architecture that has been embedded in an SSD-like network to fuse the available feature maps, giving rise to the FPSSD network. Besides, prior boxes for unoriented bounding box regression have been chosen on the basis of a clustering process over the available datasets. In the second stage, a simple but effective neural network has been designed to regress the parameters of oriented bounding boxes. The design process has considered two parameterizations of oriented bounding boxes, being the two-target RBox regression model the variant with highest performance. 

{The experimental results collected along the different evaluations performed show improved performance of FPSSD-RBox over other orientation-aware detection approaches, not only comparing favourably with them but also being able to outperform them and achieve the highest scores. At the same time, this level of performance is attained being the fastest method among the different detectors considered.}


\section*{Funding}
\label{}

This work has been partially supported by EU-H2020 projects BUGWRIGHT2 (GA 871260) and ROBINS (GA 779776), by project PGC2018-095709-B-C21 (funded by MCIU/AEI/10.13039/501100011033 and FEDER ``Una manera de hacer Europa") and by project IMABIA (PROCOE/4/2017, Govern Balear, 50\% P.O. FEDER 2014-2020 Illes Balears). This publication reflects only the authors views and the European Union is not liable for any use that may be made of the information contained therein.


\bibliographystyle{elsarticle-num} 
\bibliography{abbrv,refs1,refs2}


%
%
%
\end{document}